# Decoding Mixture Perception through Computational Modeling of Component Interactions

Fei Wang[1], Xiaoya Xie[1], Junfei Liu[1], Huihao Wang[1],Yixiao Wang[1], Yintao Wang[1], Yi Li[1], Hao Dong[2]*, Xing Chen[1]*

**Affiliations:**

[1] College of Biomedical Engineering & Instrument Science, Zhejiang University, Hangzhou, 310012, China

[2] College of Automation Engineering, Nanjing University of Aeronautics and Astronautics, Nanjing 210016, China

*Corresponding Author: Hao Dong, Xing Chen

First author email: wangfei_hangzhou@zju.edu.cn

Corresponding author email: cnhaodong@nuaa.edu.cn, cnxingchen@zju.edu.cn

This work was supported in part by the National Natural Science Foundation of China under Grant 82172064 and Grant 81571769, in part by the "Pioneer" and "Leading Goose" Research and Development Program of Zhejiang under Grant 2023C03096. Additionally, this work was supported by the National Natural Science Foundation of China (No. 62304206), the Startup Supporting Funds for Talents of NUAA to H.D., and the Open Research Project of MIIT Key Laboratory of Non-Destructive Testing and Monitoring Technology.

**Abstract:** Olfaction played an indispensable role throughout human evolution and civilization. Even in the contemporary era of advanced technology, olfaction remains a critical channel for person to conduct danger discrimination, emotional experience, and memory formation. Decoding the olfactory perception holds profound practical significance and is poised to advance the development of autonomous robotic sensing and intelligent human–machine interaction. However, most substances in nature exist as multi-molecule mixtures. The complexity of mixture compositions, nonlinear molecular interactions, as well as concentration-dependent saturation effects and receptor-specific activation thresholds, pose substantial challenges in identifying olfactory characteristics. In this study, we proposed a novel bio-inspired deep learning framework for accurate odor perception recognition of mixtures. We robustly constructed neural response curves for molecule-receptor interactions, and developed a fusion strategy that integrates attention-weighted multi-receptor curves with concentration-dependent multi-molecule curves, replicating the competitive activation and synergistic integration of mixture components. Furthermore, by comparing the consistency of response curve patterns, the model can transfer knowledge from the semantically rich space of molecular associations to guide recognition of mixture perception characteristics. Therefore, we established a complete computational pathway from chemical blending, neural encoding, to perceptual formation. Finally, we conducted comprehensive evaluation, and results demonstrated exceptional superiority, achieving an accuracy of 92.2%. Consequently, our work provides a generalizable solution to the long-standing mixture-perception challenge, effectively advancing the understanding of the olfactory mechanisms. This study is expected to deliver considerable value in disease diagnosis and industrial safety. More importantly, it can be integrated into embodied cognitive systems to enhance the agents' perceptual and interactive capabilities in complex scenarios.

## I. INTRODUCTION

Olfaction has accompanied humanity through millions of years of adaptive challenges—from hunting and foraging, hazard recognition, to reproduction, it has always been an essential sensory modality for individual survival and social cohesion [1-3]. Even in modern civilization, the perceptual function of olfaction continues to widely permeate diverse fields. In industrial production, olfactory monitoring is applied in scenarios such as food safety assurance and chemical leak early warning. In both Eastern and Western medical practices, olfaction is utilized for pathological odor analysis to assist in diagnosis [4-6]. Concurrently, with the advent of the embodied intelligence era, visual and language-based models have successfully empowered industrial applications, enabling robots to perform complex tasks. Integrating olfactory intelligence into robotic systems, granting them the ability to assess safety and danger, will substantially enhance autonomous decision-making capabilities across varied scenarios. Nevertheless, unraveling and modeling olfactory perception still faces significant challenges that need to be addressed.

Current research primarily focuses on investigating the relationship between single molecule and odor perception, which is a significant limitation, as most natural odors exist as multi-molecule mixtures rather than isolated compound [7-9]. The relationship between multi-molecule mixtures and odor perception is extraordinarily complex, exhibiting a high-dimensional interaction mapping, as shown in Figure 1. Factors such as the composition and concentrations of molecules within the mixture, alongside intricate inter-molecule interactions, and the spatiotemporal integration of receptor activation collectively form a multi-level, cross-scale perceptual coding network that can significantly shape the final odor perception. Consequently, the olfactory decoding of mixtures involves not only a combinatorial explosion in chemical space but also engages fundamental neural mechanisms such as receptor competition and perceptual emergence [10-12]. Breaking through this bottleneck would enable us to construct interpretable intelligent perception system, and empower robotic systems to make more autonomous decisions in complex scenarios.

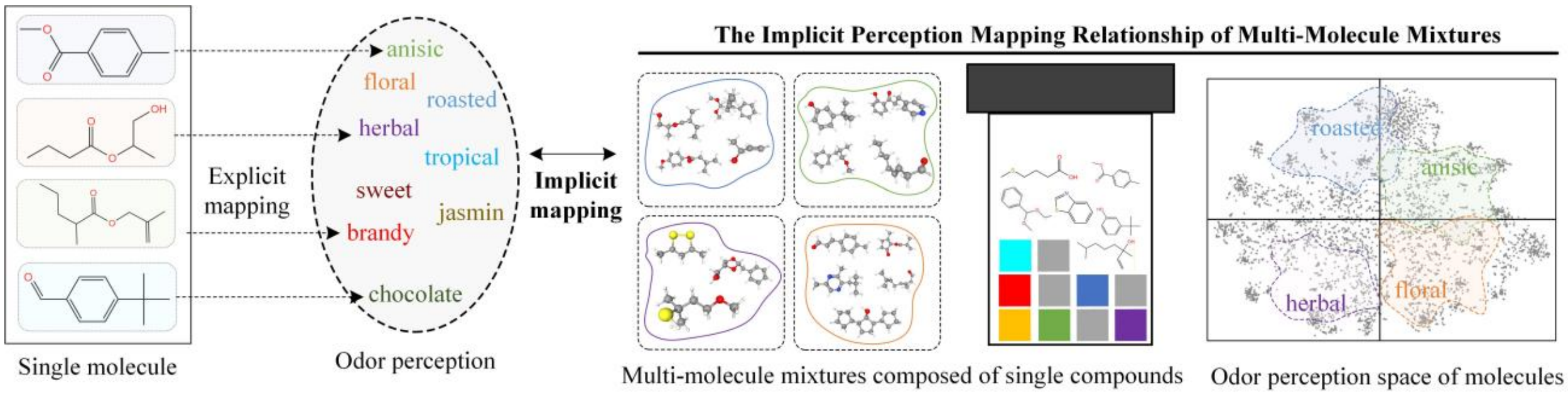

**Figure 1. Mapping between Molecule and Odor Perception.** The explicit mapping between single molecule and odor perception (left). The composition of multi-molecule mixtures is complex, with varying types and concentrations of molecules, which significantly influence odor perception, leading to an implicit mapping (right).

The objective of our study is to accurately identify the odor perception of multi-molecule mixtures, clarify the core transduction mechanisms of multi-molecule mixtures within the biological olfactory system and to establish a complete computational pathway from chemical blending to neural encoding and perceptual formation. To achieve this, we innovatively proposed a deep-learning framework that incorporates bio-inspired olfactory principles for accurate odor-perception identification. However, due to the lack of sufficient mixture-perception datasets for model training, we did not adopt an end-to-end mixture-perception mapping approach. Instead, we leveraged the abundant existing single-molecule perception data to guide the prediction of mixture perception by establishing semantic associations between molecule and mixture response patterns. This enables the model to transfer knowledge from the rich semantic space of single molecules, achieving robust recognition of complex mixture odor characteristics with limited labeled data. Specifically, we firstly designed a molecule-to-receptor neural response curve prediction model, which can accurately characterize the response curves arising from their interactions. Furthermore, recognizing that molecule concentration significantly influences odor perception, we developed a mixture multi-receptor response curve with concentration-dependent multi-molecule curves fusion strategy, to simulate the competitive activation and synergistic integration of mixture components. The fusion strategy effectively reflects the impact of the composition and concentration of individual molecules within the mixture on the final odor perception. Subsequently, the model is optimized by minimizing the discrepancy between the odor perception of molecules and response curves. Finally, we utilized single-molecule odor perception data as a prior, guiding the identification of mixture odor perception. By comparing the consistency of single-molecule and multi-molecule response curves and dynamically assigning differentiated weights to odor features based on curve pattern similarity, the odor perception of multi-molecule mixtures can be accurately recognized. The overall workflow is illustrated in Figure 2 (Step 1 to 5). The detailed descriptions and formulation description of each step are provided in Appendix II and III, specifically.

We conducted a series of comprehensive experiments to evaluate the performance of our method. First, we compared the predicted molecule–receptor response curves with ground-truth recorded curves to validate the accuracy of our response curve modeling (Section II-A). Second, we examined whether the discrepancies between neural response curves are consistent with the differences in odor perception, thereby verifying the proposed isomorphic relationship (Section II-B). Finally, we compared our approach with existing perception similarity recognition methods for mixtures (Section II-C), as well as with other deep learning models for odor perception identification of mixtures (Section II-D), further demonstrating the superior performance of proposed method. The experimental results demonstrated that our method achieved significant improvements in performance metrics, reaching an accuracy of 92.2% on a newly curated real-world dataset. Our method can accurately identify the odor perception of mixtures. Thus, our work elucidates the formation of olfactory perception in mixtures, offering a computationally tractable solution to the long-standing challenge of chemical blending to percept emergence in olfactory science.

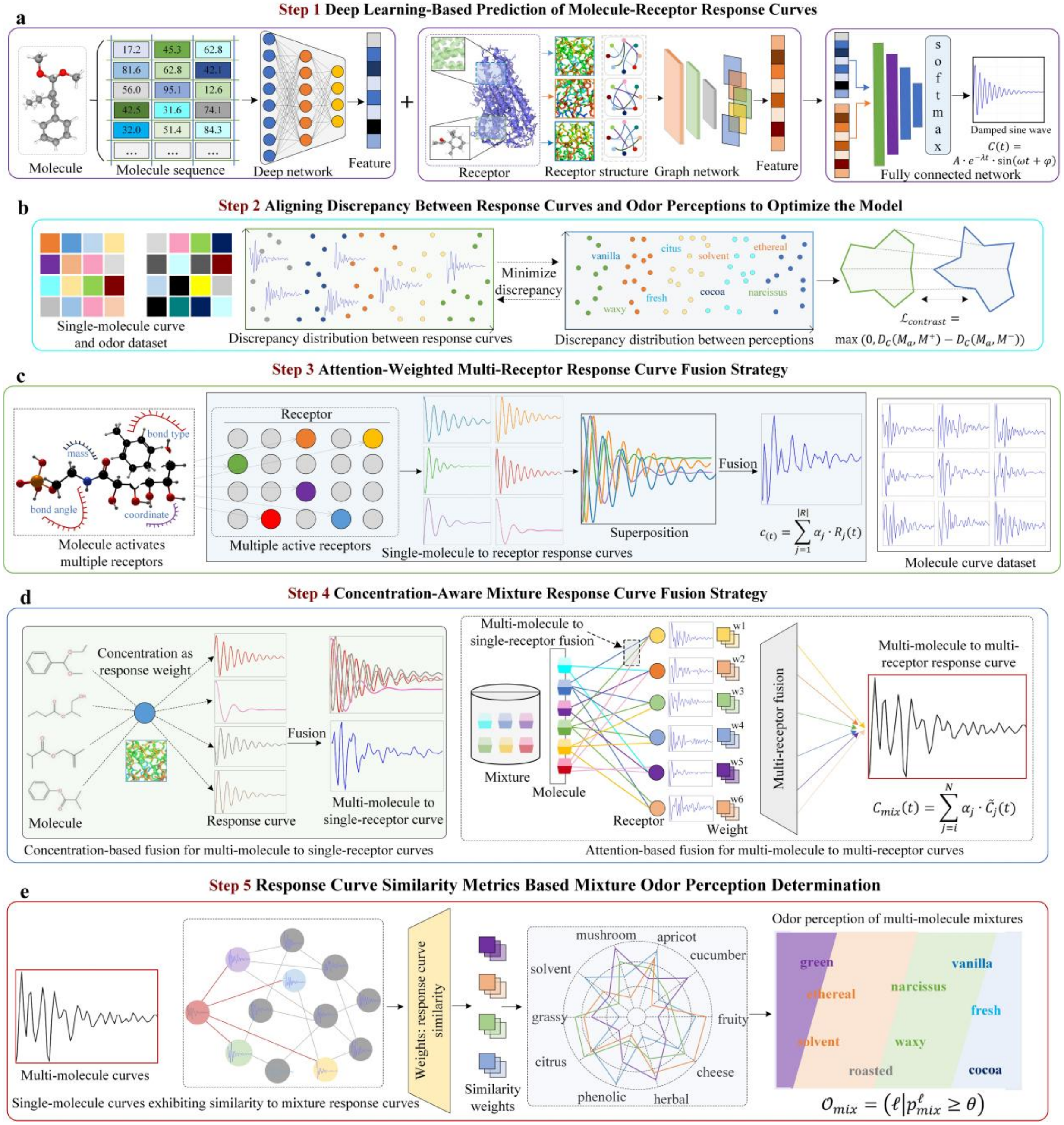

**Figure 2. Overall Workflow of the Proposed Method.** Firstly, we developed a deep learning-based molecule to receptor response curve prediction model. Secondly, we leveraged the consistency between molecule odor perception and response curves, optimizing the model by minimizing and aligning the discrepancies between the two, enabling the model to learn a stable mapping relationship between perception and response curves. Thirdly, based on biological response principles, we established an attention-weighted multi-receptor response curve fusion strategy, thus constructing response curves at the multi-receptor level. Fourthly, recognizing that molecule concentration significantly influences odor perception, we designed a concentration-aware mixture response curve fusion strategy. Fifthly, we utilized single-molecule odor perception data as a prior, guiding the identification of mixture odor perception, thereby achieving an accurate mapping from the neural response domain to the odor perception space.

## II. Experiments

### A. Robustness Validation of Neural Response Curves

To validate the accuracy and biological plausibility of the predicted molecule-receptor neural response curves, we collected real neural response curve data induced by multi-molecule mixtures from the existing experiments. These mixtures consist of various single molecules with significant differences in concentration ratios, thereby reflecting the complexity of real olfactory stimulus environments. Ultimately, we assembled a dataset of 35 mixture-response curve pairs (detailed collection and curation procedures are provided in Appendix I-E) for external validation. All data were derived from biochemical experiments and represent dynamic biosignal responses generated by interactions between multi-molecule odor mixtures and olfactory receptors.

We assessed the closeness between the real neural response curves and the predicted curves using curve similarity. Curve similarity is defined as the degree of similarity between the response values of the two curves at corresponding time points. The experimental results show that the matching score of our method is 0.925, indicating minimal deviation with high consistency. Moreover, we compared our method with other models, including traditional models (RF, CNN) [13, 14] benchmark models (DKNN, GNN)[15, 16], and advanced pre-trained molecular models (POM, MolFormer)[17, 18]. Figure 3a presents the response curves predicted by each model, demonstrating that our method achieves the closest alignment with the ground-truth curves. As shown in Figure 3b, the curve similarities for RF and CNN were 0.712 and 0.732, respectively; for DKNN and GNN, the curve similarities were 0.765 and 0.802; for POM and MolFormer, the curve similarities were 0.835 and 0.864. In addition, we evaluated whether the response peaks were aligned to assess method performance, i.e., the degree of positional matching between the predicted response peaks and the real response peaks. Our method achieved a matching result of 0.945, while the results obtained by other methods were all below 0.850. This indicates that our method can accurately predict the peaks of the curves. Figure 3c shows the matching accuracy for each of the 35 mixture cases obtained by different methods, where our method achieves the highest performance across all cases. Compared to other six models, our method significantly outperformed them in terms of response curve prediction accuracy. The experimental results demonstrate that our method can accurately reconstruct neural response curves under complex, diverse real mixture conditions, maintaining stable and superior performance on external datasets, thereby confirming the robustness, practicality, and strong generalization capability of our method.

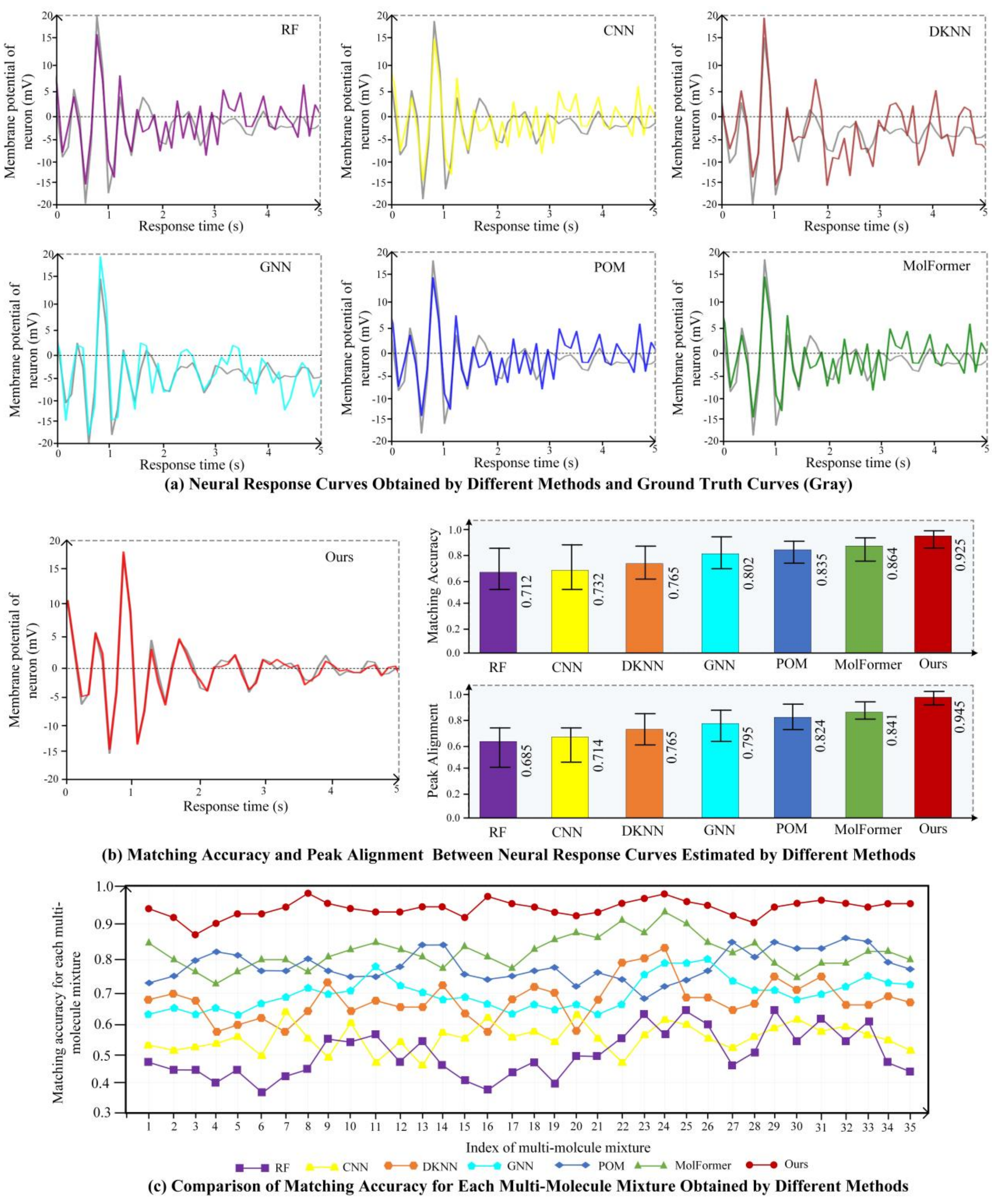

**Figure 3. The Response Curves Obtained by Different Methods and Ground Truth Curves.** (a) Neural response curves predicted by different methods. The results obtained by our method show the closest alignment with the ground-truth curves. (b) Comparison of matching accuracy and peak alignment of response curves obtained by different methods. (c) Matching accuracy for each multi-molecule mixture obtained by different methods. Our method achieved the highest accuracy.

### B. Performance of Discrepancy Alignment Between Response Curves and Odor Perceptions

In this section, we validated the isomorphic relationship between neural response curves and odor perception differences. Specifically, this means that when perceptual differences between molecules are large, their corresponding neural response curve differences are also substantial, and vice versa. Using a dataset of 5,023 molecules (detailed information is provided in Appendix I-A), we performed pairwise comparisons among all molecules and constructed a complete set of molecule pairs, resulting in approximately 12.6 million sample pairs. For each molecule pair, we predicted the neural response curves induced by the two individual molecules using our method. We then calculated the pairwise response curve similarity and the corresponding pairwise odor perception similarity for the same molecule pair. Finally, we used the Pearson correlation coefficient (r) to quantify the global alignment between these two similarity metrics across all molecule pairs, which allows us to evaluate whether variations in response curve similarity are proportionally reflected in perceptual similarity. This metric serves as a quantitative indicator of whether the learned neural response representation preserves the geometric structure of the odor perception space.

The correlation was found to be $r = 0.921$. This significant positive correlation indicates that our model successfully captured the topological structure of the perceptual space: the greater the odor perception difference between molecules, the larger the difference in the neural response curves predicted by the model. As shown in Figure 4a, the density plot illustrates a linear positive correlation and isomorphic relationship between the response curve and odor perception differences. Additionally, we evaluated the performance of other methods in assessing the isomorphic relationship between curves and perceptual similarity. The correlations obtained were RF (0.521), CNN (0.558), DKNN (0.635), GNN (0.721), POM (0.815), and MolFormer (0.821), all of which were substantially inferior to the performance of our method. Therefore, our approach demonstrates superior discriminative power, effectively aligning the differences in molecule response curves with odor perception differences.

In Figure 4b, we selected 70 of the most challenging and representative molecules, which capture the main characteristics of the molecule dataset. The criteria for molecule selection are as follows. Functional group diversity: The molecules in the triplets cover common functional groups related to perception, including carboxyl, hydroxyl, amino, and ether groups. Chemical category diversity: The selected molecules cover major chemical categories, including hydrocar-bons, alcohols, esters, and aromatic compounds. Molecule structure diversity: The selected molecules vary in molecule weight (from simple low-weight compounds to complex high-weight compounds) and shape (linear, cyclic, and branched). Similarly, for each representative molecule, we computed pairwise similarities between its predicted response curve and those of all other molecules, as well as the corresponding pairwise odor perception similarities. The correlation was then calculated between these two similarity distributions for each representative molecule. Our method achieved the optimal correlation coefficients across all these molecules, with all metric values remaining above 0.82 (Figure 4b, red points), indicating that the predicted response curves are highly consistent with their corresponding odor perceptions. In contrast, the performance of the other six comparative methods was markedly inferior.

In Figure 4c, we further analyzed how the number of odor perceptions associated with the representative molecules influenced the model's performance. For these representative molecules, the number of associated odor perceptions ranged from 1 to 15, meaning that each molecule was linked to at least one odor perception and at most fifteen odor perceptions in the dataset. The upper bound was determined by the dataset itself, as the maximum number of odor perception associated with any single molecule was 15. The results show that across this range, from molecules associated with only one odor perception to those associated with up to fifteen odor perceptions, the correlation results obtained by our method remained consistently high, demonstrating strong stability across molecules with different levels of perceptual complexity. This indicates that even when the odor perceptions corresponding to a molecule are complex and diverse, our method can still accurately predict the response curves elicited by that molecule. The gray points in the figure represent the average results of the other six models, whose results gradually declined as the number of odor perceptions increased, reflecting poorer predictive performance.

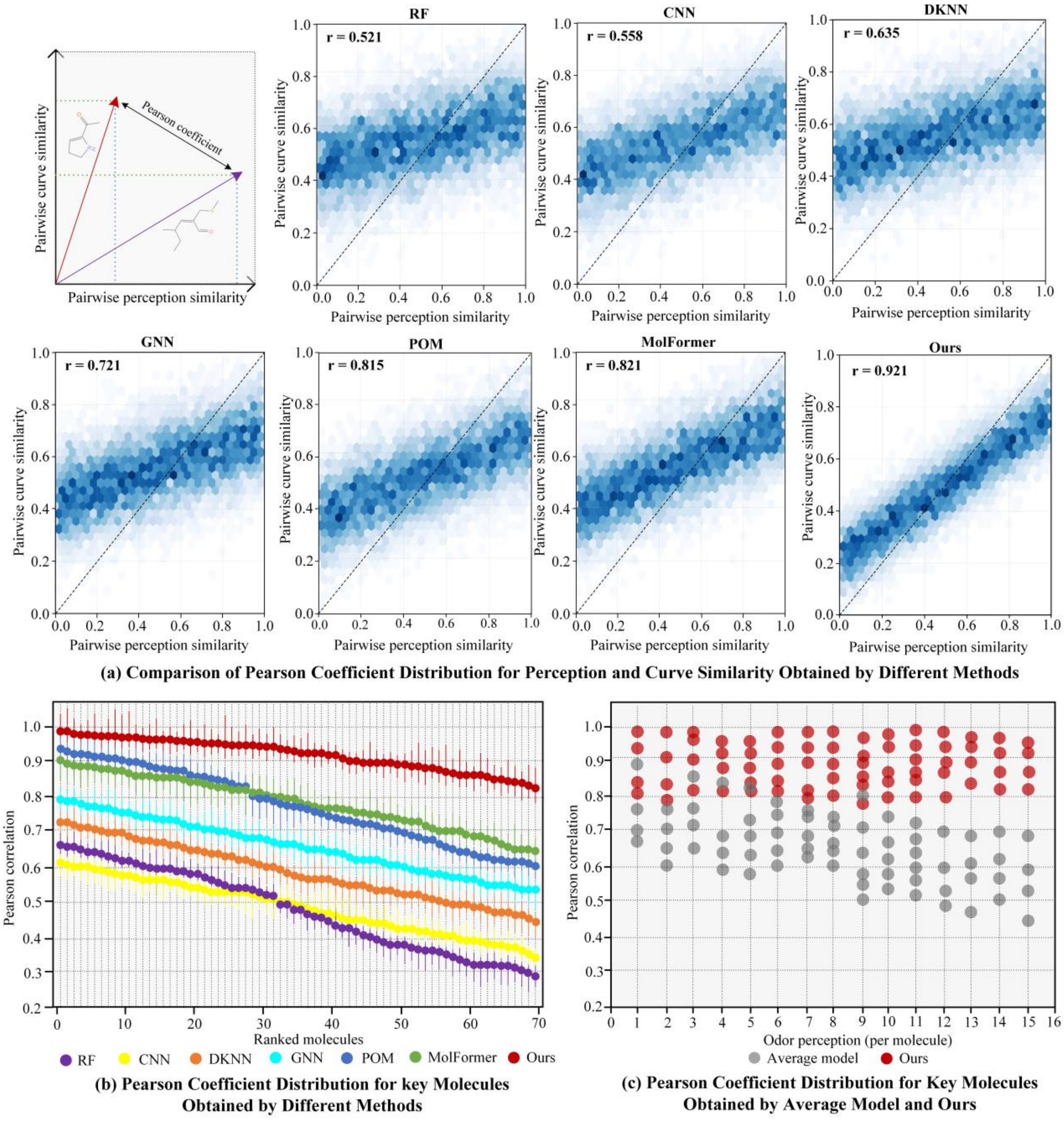

**Figure 4. Correlation Comparison between Molecular Odor Perception and Response Curves Obtained by Different Methods**

### C. Evaluation of Odor Perception Similarity for Mixtures

To evaluate the ability of our method to recognize odor perception in real-world multi-molecule mixtures, we curated and constructed a real dataset containing 112 mixtures (detailed description provided in Appendix I-C). This dataset provides detailed annotations of the individual molecule components in each mixture, along with the ground-truth perceptual similarity scores between each pair of mixtures, obtained through human sensory testing. We identified the odor perception for each mixture using our method, and quantified the similarity in odor perception between mixtures by calculating cosine similarity. Finally, we applied the Pearson correlation coefficient to assess whether the odor perception similarities calculated by our method align with the ground-truth human perceptual similarities. Additionally, we used the Mean Absolute Error (MAE) to evaluate the absolute deviation between two similarities, that is, the difference between the perceptual similarity among mixtures calculated by our method and the ground-truth perceptual similarity. The MAE ranges from [0, 1], where values closer to 0 indicate smaller deviations and better model performance.

Furthermore, we compared our method with two existing recognition models, namely Amit et al. [19] and Ravia et al. [20]. Both models are specifically designed to identify perceptual similarity between molecule mixtures. They predict odor perception from molecular structures, quantify odor similarity by measuring the angular distance between molecular vectors, and incorporate molecular concentration as a weighting factor to account for concentration-dependent effects. This enables the estimation of perception similarity of mixtures. Detailed descriptions are provided in Appendix V-B. The correlation between the odor perception similarity of the mixtures calculated by our method and the ground-truth similarity is $r = 0.931$. The result obtained using Amit et al. is 0.811, and obtained using Ravia et al. is 0.834, both of which are significantly inferior to the result obtained by our method. The distribution of perceptual similarity for the mixtures is shown in Figure 5a, where our method achieves better correlation. Additionally, the MAE (Figure 5b) obtained by our method is 0.072, while these derived from the Amit et al. and Ravia et al. are 0.352 and 0.224, respectively. Our approach yields a lower MAE, indicating improved accuracy. Meanwhile, our method achieves a Spearman value of 0.962 (an improvement of 0.211 and 0.200 compared to Amit et al. and Ravia et al.), an RMSE of 0.084 (an improvement of 0.089 and 0.081), an F1 score of 0.945 (an improvement of 0.141 and 0.110), and an $R^2$ of 0.942 (an improvement of 0.166 and 0.127). These results demonstrate significant improvements over both Amit et al. and Ravia et al.

Additionally, we compared our method with two advanced models, POM and MolFormer (see Appendix V-B). These methods represent state-of-the-art approaches in olfactory perception prediction. For these models, odor perception of molecule was first predicted from chemical structures. The predicted molecule perceptions were then integrated using the concentration-weighting strategy adopted in this study to infer mixture perception, after which perceptual similarity between mixtures was evaluated. These models yielded correlations of 0.721 and 0.755, and MAE values of 0.415 and 0.381, respectively, both of which were substantially inferior to those achieved by our method (Figure 5c). Therefore, our method accurately aligns the odor perception similarity with the ground-truth similarity. This low MAE further demonstrates that our method can accurately distinguish the odor perception differences between mixtures.

In Figure 5d, we selected the 16 most frequently occurring odor perceptions in the dataset and presented the perceptual distributions of multi-molecule mixtures predicted by different methods, comparing them with the ground-truth odor perception distributions. The matching accuracies obtained by each method are as follows: POM (0.715), MolFormer (0.772), Amit et al. (0.795),

Ravia et al. (0.836), and our method (0.942). The perceptual distribution obtained by our method aligns most closely with the ground-truth results, while those obtained by other methods exhibit considerable deviations.

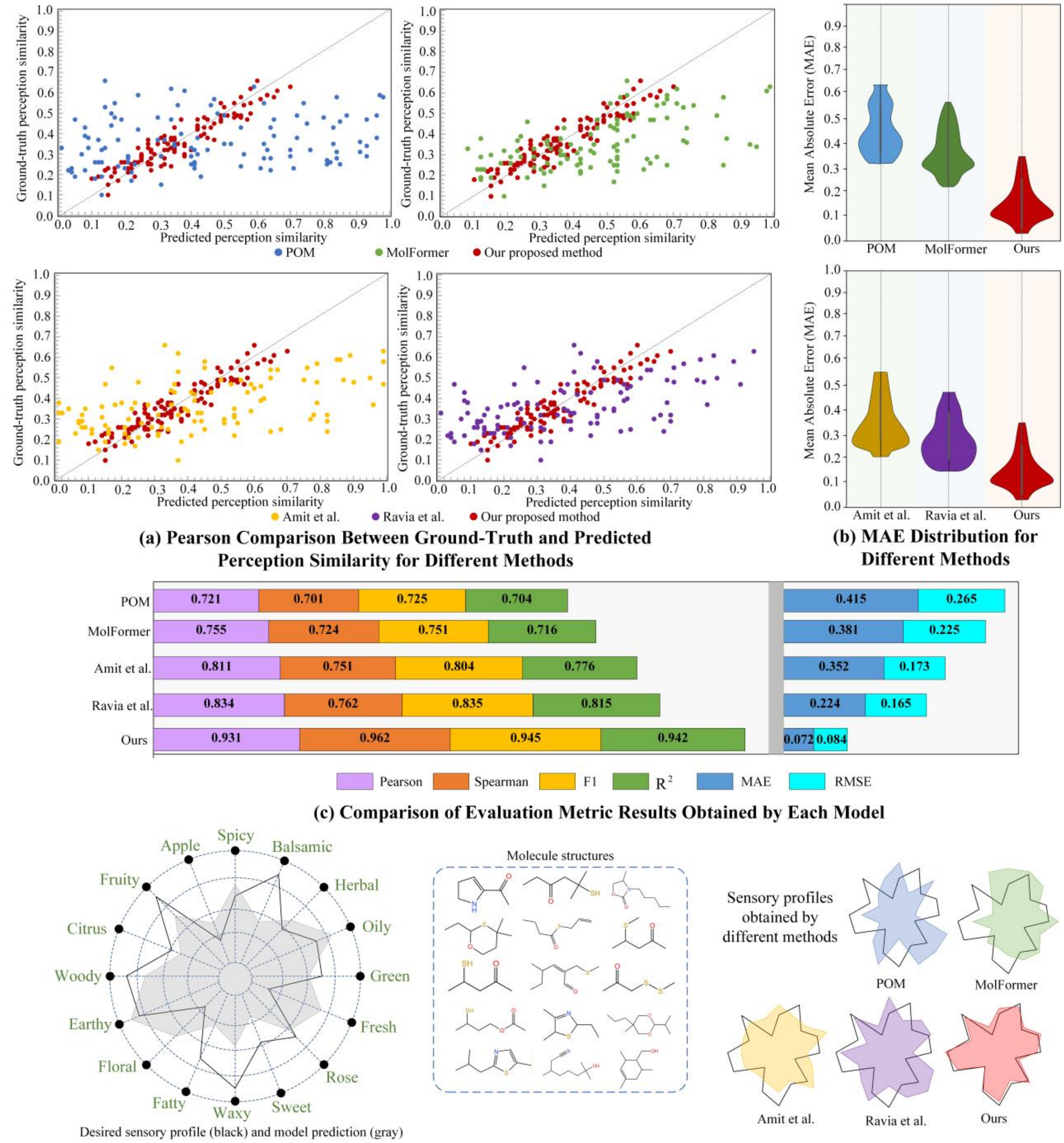

**Figure 5. Comparison of Odor Perception Similarity Results for Multi-Molecule Mixtures Obtained by Different Methods**

## D. Evaluation of Odor Perception Recognition for Mixtures

To assess the robustness of our method in recognizing the specifical odor perception of real-world mixtures, we collected a dataset containing odor perception data for 172 mixtures. This dataset is highly challenging, as the mixtures feature complex chemical compositions, including a diverse array of single molecule components and precise concentration ratios, offering detailed odor perceptions for raw materials and natural mixtures (detailed description in Appendix I-D). For all baseline methods, we adopted the same strategy to obtain odor perceptions of mixture. Specifically, each model first predicted the odor perception of individual molecules within a mixture based on molecular representations via an end-to-end mapping strategy. The mixture perception was then derived by aggregating these single-molecule predictions using a concentration-dependent weighting scheme in our framework, where molecules with higher concentrations contribute more strongly to the final perception. Consequently, the overall odor perception of a mixture can be inferred from constituent molecular predictions.

We used cosine similarity as a quantification metric to calculate the alignment between the predicted odor perceptions and the ground-truth human odor perceptions. The cosine similarity ranges from 0 to 1, with values closer to 1 indicating better model performance. Furthermore, we compared our method with traditional classification models (RF and CNN), benchmark models (DKNN and GNN), and advanced pre-trained molecule models (POM and MolFormer). Firstly, we evaluated the cosine value between the odor perception of the mixtures calculated by different models and the ground-truth odor perception. The cosine value obtained by our method was 0.942, while RF and CNN resulted in 0.657 and 0.661, respectively. The results from DKNN and GNN were 0.742 and 0.774, significantly worse than those computed by our method. The results from POM and MolFormer were 0.814 and 0.832. Therefore, the evaluation based on cosine similarity demonstrates that our method significantly outperforms other models and can accurately identify the odor perception corresponding to multi-molecule mixtures.

**Table 1. Performance Comparison of Different Methods.**

| Model | AUROC | AUPRC | Precision | Recall | Specificity | Accuracy |
|---|---|---|---|---|---|---|
| RF | 0.682 ± 0.203 | 0.648 ± 0.225 | 0.735 ± 0.215 | 0.714 ± 0.124 | 0.664 ± 0.261 | 0.665 ± 0.195 |
| CNN | 0.634 ± 0.214 | 0.674 ± 0.203 | 0.671 ± 0.135 | 0.675 ± 0.151 | 0.712 ± 0.135 | 0.684 ± 0.224 |
| DKNN | 0.745 ± 0.141 | 0.751 ± 0.211 | 0.826 ± 0.152 | 0.862 ± 0.141 | 0.761 ± 0.184 | 0.725 ± 0.115 |
| GNN | 0.835 ± 0.084 | 0.724 ± 0.152 | 0.834 ± 0.126 | 0.824 ± 0.062 | 0.824 ± 0.138 | 0.764 ± 0.084 |
| POM | 0.844 ± 0.062 | 0.836 ± 0.116 | 0.822 ± 0.127 | 0.847 ± 0.085 | 0.834 ± 0.092 | 0.802 ± 0.062 |
| MolFormer | 0.851 ± 0.051 | 0.852 ± 0.048 | 0.841 ± 0.075 | 0.875 ± 0.074 | 0.872 ± 0.074 | 0.826 ± 0.054 |
| **Ours** | **0.924 ± 0.031** | **0.919 ± 0.018** | **0.932 ± 0.028** | **0.916 ± 0.052** | **0.915 ± 0.021** | **0.922 ± 0.016** |

Table 1 presents the results of the six evaluation metrics in detail. Figure 6a further illustrates the distribution of each model's performance across the metrics using box plots, while Figure 6b depicts the ROC curves for different models and our method. The experimental results demonstrate the exceptional performance of our method, with the following metrics: AUROC of 0.924 ± 0.031, AUPRC of 0.919 ± 0.018, Precision of 0.932 ± 0.028, Recall of 0.916 ± 0.052, Specificity of 0.915 ± 0.021, and Accuracy 0.922 ± 0.016. Notably, the small standard deviations of these metrics provide strong evidence for the outstanding stability and generalization capability in odor perception prediction tasks. Subsequently, we compared our method with the advanced models POM and MolFormer. Although POM achieved AUROC of 0.844, AUPRC of 0.836, and accuracy of 0.802, our method significantly surpassed these results across all metrics, improving by 0.080, 0.093, and 0.120, respectively. Compared to MolFormer, our method also achieved performance gains of 0.073, 0.067, and 0.096 in the same three metrics. Notably, our method showed particularly impressive improvements in recall and specificity: recall increased from 0.847 in POM and 0.875 in MolFormer to 0.916, while specificity significantly improved from 0.834 and 0.872 to 0.915, respectively.

Figure 6c presents the heatmap distribution of the 16 most common odor perceptions. Figure 6d presents the connection strengths among these most common odors, where the connection strength is calculated based on the number of molecules shared between odors. The results reveal that these most common odors exhibit significant correlations with other odors, sharing a substantial number of common molecules. This suggests that predicting frequent odors in mixtures is a highly challenging task, yet our method maintains high accuracy. In Figure 6e, we further evaluated and ranked the recognition accuracy of common odor perceptions. Specifically, after calculating the perception recognition accuracy for each mixture, we further performed an odor-category-level analysis. For each odor perception, we calculated its recognition accuracy across all relevant mixture, thereby evaluating the model's ability to identify different odor perceptions. The results show that all accuracy values are above 0.9, indicating that our method can reliably identify these odor perceptions corresponding to the mixtures.

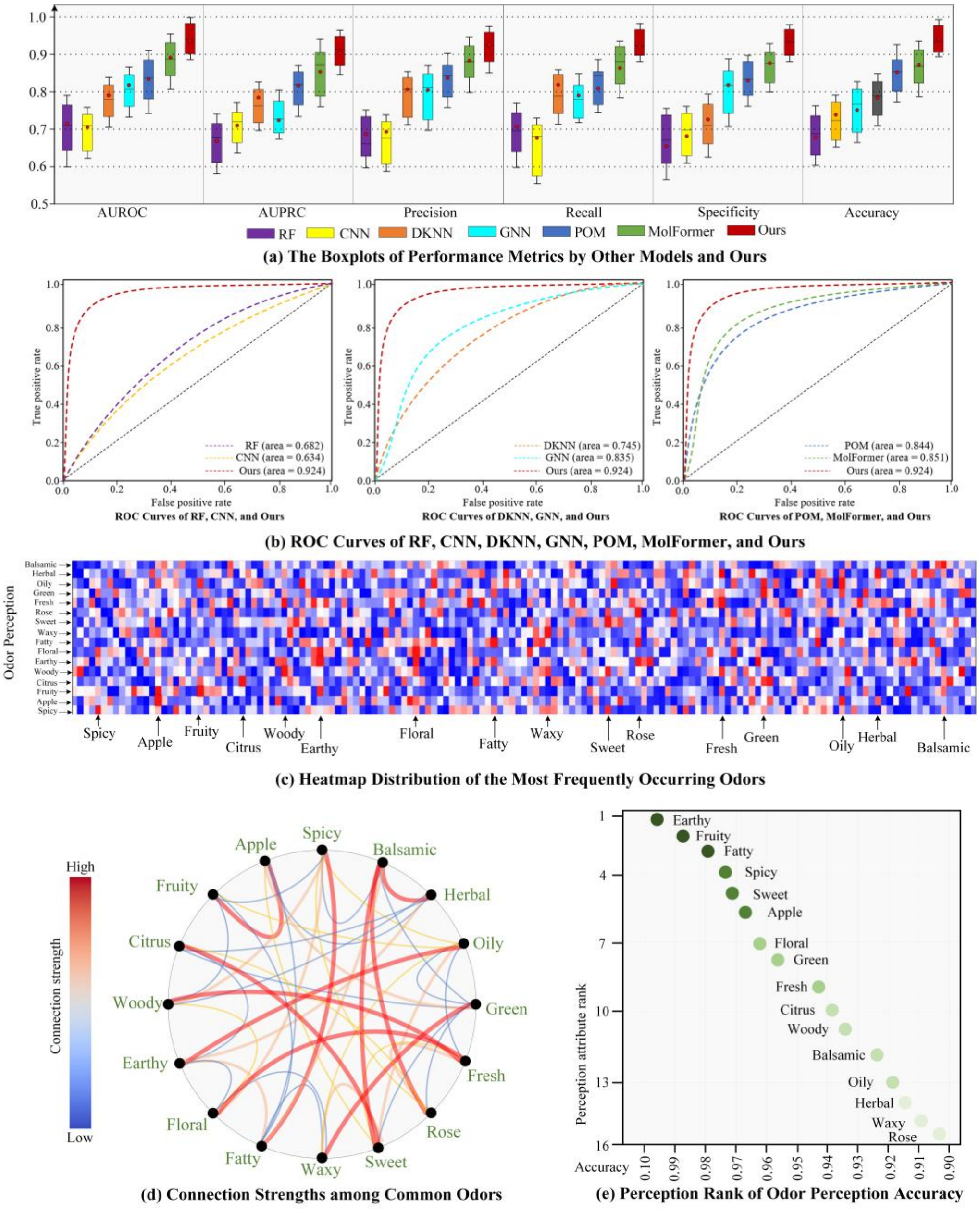

**Figure 6. Performance Comparison of Odor Perception Prediction for Multi-Molecule Mixtures Obtained by Different Models**

## III. Discussion

In this paper, we addressed the long-standing challenge of decoding olfactory recognition in multi-molecular mixtures—a fundamental problem in human sensory science—by developing a biologically grounded olfactory coding model. We presented a novel approach for identifying the odor perception of multi-molecule mixtures. First, we designed the deep learning model to characterize the topological structural features of molecules and olfactory receptors, and constructed neural response curves to represent their interactions using bio-inspired mechanisms. Furthermore, we proposed an attention-weighted multi-receptor response curve fusion strategy with a concentration-aware mixture response curve fusion scheme to reflect the influence of different receptors and molecule concentrations on odor perception. Additionally, due to the scarcity of real-world data linking multi-molecule mixtures to odor perception, we devise an unsupervised pretraining paradigm based on contrastive learning. Finally, a transfer-learning strategy is adopted to leverage extensive single-molecule perception data as prior knowledge, guiding the inference of complex mixture perception. To validate the performance of proposed method, we collected a real-world multi-molecule mixture dataset containing different molecules and concentration ratios, and through comparisons with traditional models, benchmark models, and advanced models, we verified the robustness of our method. The experimental results show that our method achieves an accuracy of 92.2% and can accurately identify the odor perceptions of multi-molecule mixtures. Therefore, the proposed method offers a high-precision computational framework for olfactory perception recognition and establishes an interpretable pathway from chemical stimulation to neural encoding and, ultimately, to perceptual formation. Our work lays a technical foundation for the digital modeling of the olfactory system and holds promising potential for future deployment in embodied scenarios.

## References

[1] A. Keller, R. C. Gerkin, Y. Guan, A. Dhurandhar, G. Turu, B. Szalai, J. D. Mainland, Y. Ihara, C. W. Yu, R. Wolfinger. Predicting human olfactory perception from chemical features of odor molecules [J]. Science, 2017, 355(6327): 820-6.

[2] K. A. Fulton, D. Zimmerman, A. Samuel, K. Vogt, S. R. Datta. Common principles for odour coding across vertebrates and invertebrates [J]. Nature Reviews Neuroscience, 2024, 25(7): 453-72.

[3] G. Bratman, C. Bembibre, G. Daily, R. Doty, T. Hummel, L. Jacobs, P. Kahn Jr, C. Lashus, A. Majid, J. Miller. Nature and human well-being: the olfactory pathway. Sci Adv 10: eadn3028 [Z]. 2024

[4] M. Zhang, L. Zhu, J. He, Y. Liu, S. Ding, X. Lin. Clinical study on the application of a high-sensitivity electronic nose on thin-film gas sensor array technology combined with deep learning algorithm for early non-invasive diagnosis of chronic atrophic gastritis [J]. Biomedical Signal Processing and Control, 2025, 107: 107851.

[5] L. Aziz, H. Adil, R. Sarwar. Artificial sensing: AI-driven electronic nose for real-time gas leak detection and food spoilage monitoring [J]. Sir Syed University Research Journal of Engineering & Technology, 2025, 15(1): 71-82.

[6] M. Aleixandre, D. Prasetyawan, T. Nakamoto. Automatic scent creation by cheminformatics method [J]. Scientific Reports, 2024, 14(1): 31284.

[7] L. Xu, W. Li, V. Voleti, D.-J. Zou, E. M. Hillman, S. Firestein. Widespread receptor-driven modulation in peripheral olfactory coding [J]. Science, 2020, 368(6487): eaaz5390.

[8] K. Snitz, A. Yablonka, T. Weiss, I. Frumin, R. M. Khan, N. Sobel. Predicting odor perceptual similarity from odor structure [J]. PLoS computational biology, 2013, 9(9): e1003184.

[9] E. A. Hamel, J. B. Castro, T. J. Gould, R. Pellegrino, Z. Liang, L. A. Coleman, F. Patel, D. S. Wallace, T. Bhatnagar, J. D. Mainland. Pyrfume: A window to the world's olfactory data [J]. Scientific data, 2024, 11(1): 1220.

[10] G. Tom, C. T. Ser, E. M. Rajaonson, S. Lo, H. S. Park, B. K. Lee, B. Sanchez-Lengeling. From Molecules to Mixtures: Learning Representations of Olfactory Mixture Similarity using Inductive Biases [J]. arXiv preprint arXiv:250116271, 2025.

[11] C. S. Sell. On the unpredictability of odor [J]. Angewandte Chemie International Edition, 2006, 45(38): 6254-61.

[12] C. Bushdid, M. O. Magnasco, L. B. Vosshall, A. Keller. Humans can discriminate more than 1 trillion olfactory stimuli [J]. Science, 2014, 343(6177): 1370-2.

[13] H. A. Salman, A. Kalakech, A. Steiti. Random forest algorithm overview [J]. Babylonian Journal of Machine Learning, 2024, 2024: 69-79.

[14] Z. Li, F. Liu, W. Yang, S. Peng, J. Zhou. A survey of convolutional neural networks: analysis, applications, and prospects [J]. IEEE transactions on neural networks and learning systems, 2021, 33(12): 6999-7019.

[15] N. Papernot, P. Mcdaniel. Deep k-nearest neighbors: Towards confident, interpretable and robust deep learning [J]. arXiv preprint arXiv:180304765, 2018.

[16] W. Shi, R. Rajkumar. Point-gnn: Graph neural network for 3d object detection in a point cloud; proceedings of the Proceedings of the IEEE/CVF conference on computer vision and pattern recognition, F, 2020 [C].

[17] B. K. Lee, E. J. Mayhew, B. Sanchez-Lengeling, J. N. Wei, W. W. Qian, K. A. Little, M. Andres, B. B. Nguyen, T. Moloy, J. Yasonik. A principal odor map unifies diverse tasks in olfactory perception [J]. Science, 2023, 381(6661): 999-1006.

[18] J. Ross, B. Belgodere, V. Chenthamarakshan, I. Padhi, Y. Mroueh, P. Das. Large-scale chemical language representations capture molecular structure and properties [J]. Nature Machine Intelligence, 2022, 4(12): 1256-64.
[19] A. Dhurandhar, H. Li, G. A. Cecchi, P. Meyer. Expansive linguistic representations to predict interpretable odor mixture discriminability [J]. Chemical senses, 2023, 48: bjad018.
[20] A. Ravia, K. Snitz, D. Honigstein, M. Finkel, R. Zirler, O. Perl, L. Secundo, C. Laudamiel, D. Harel, N. Sobel. A measure of smell enables the creation of olfactory metamers [J]. Nature, 2020, 588(7836): 118-23.

# Appendix

## Decoding Mixture Perception through Computational Modeling of Component Interactions

Fei Wang[1], Xiaoya Xie[1], Junfei Liu[1], Huihao Wang[1],Yixiao Wang[1], Yintao Wang[1], Yi Li[1], Hao Dong[2]*, Xing Chen[1]*

**Affiliations:**

[1] College of Biomedical Engineering & Instrument Science, Zhejiang University, Hangzhou, 310012, China

[2] College of Automation Engineering, Nanjing University of Aeronautics and Astronautics, Nanjing 210016, China

*Corresponding Author: Hao Dong, Xing Chen

First author email: wangfei_hangzhou@zju.edu.cn

Corresponding author email: cnhaodong@nuaa.edu.cn, cnxingchen@zju.edu.cn

### I. Materials and Datasets

Due to the scarcity of extensive multi-molecule mixture perception datasets, directly using these data for deep learning model training is not feasible. However, numerous single-molecule perception datasets exist. Our approach utilizes single-molecule perception to guide the identification of multi-molecule perception. Therefore, we collected single-molecule receptor datasets and single-molecule perception datasets. Additionally, to evaluate the performance of our method, we gathered the real-world mixture similarity rating dataset and mixture–perception dataset. Descriptions of each dataset are provided below.

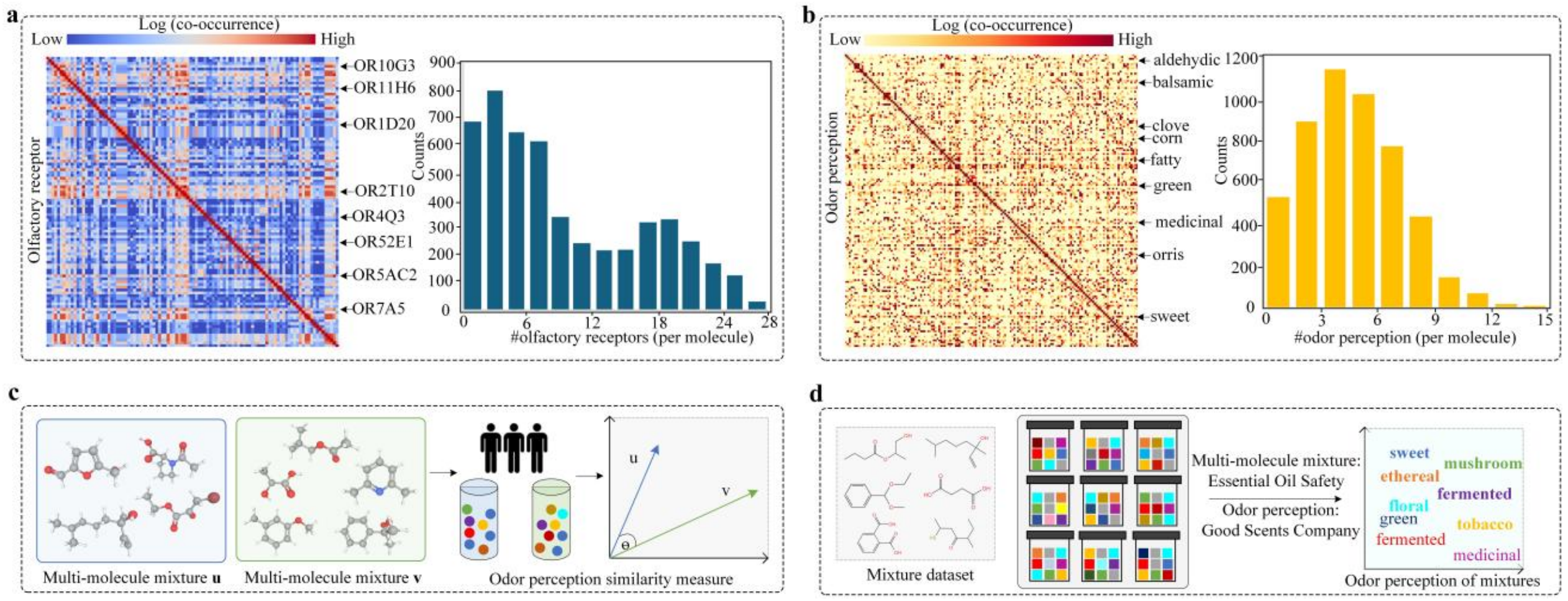

**Fig. S1. Descriptions of Each Dataset. a.** Feature distribution of molecule-receptor dataset: The distribution of molecules associated with each receptor is imbalanced; 18 receptors are linked to 100–200 molecules, while 7 receptors are associated with 500–800 molecules. **b.** Feature distribution of molecule-odor dataset: This distribution is also skewed, with 2,160 molecules having 3-4 odors, and 672 molecules possessing 6-10 odors. Additionally, the frequency of different odors varies. **c.** Mixture similarity rating dataset: This dataset contains odor perception similarity ratings between pairs of mixtures, which were evaluated through human olfactory psychophysical experiments. **d.** Mixture-perception dataset: We collected mixtures from the real world along with corresponding odor perceptions, including specific chemical compositions and concentration ratios of molecules within the mixture, for directly validating the performance of our method.

### A. Molecule-Receptor Dataset

To investigate the interaction mechanisms between odorant molecules and olfactory receptors, an essential section of olfactory perception, we collected a single-molecule receptor dataset. The dataset includes single molecules and corresponding responsive receptors. A detailed description of the dataset is provided below. The feature distribution of molecule-receptor dataset is shown in Figure S1 (a).

The molecule-receptor relationships are sourced from various datasets, including ODORDB, ODORactor, and OlfactionDB [1-2], which are widely used and reliable resources validated through real-world studies. In addition, we referred to the compiled data from other studies [3], which provide a substantial amount of molecule-receptor interaction information. They provided detailed descriptions of the receptor characteristics of each molecule compound and olfactory receptors are labeled with one or more molecules. The correspondences in the dataset are primarily derived from published academic articles, documenting the interactions between specific molecules and olfactory receptors. We manually collected the data, merged and filtered the molecule-receptor relationships from multiple databases, and eliminated duplicate entries to ensure the uniqueness of each data point. Furthermore, the sample data containing missing values and erroneous labels were directly excluded to prevent biases, which improved the overall data quality. Redundant molecule-receptor pairs were removed based on unique identifiers. To ensure data reliability, we cross-checked experimental conditions and quantitative parameters. In cases of discrepancies between datasets, we conducted thorough literature reviews to resolve conflicts. These integration strategies effectively mitigate data biases, and the obtained correspondence between molecules and receptors is validated and authentic.

In total, the curated dataset integrates 5,023 distinct single molecules with 104 olfactory receptors, ensuring high data fidelity and reliability. This dataset is employed for the subsequent biologically grounded modeling of mixture olfactory perception within the proposed framework. The dataset is provided as Supplementary Material 1.

### B. Molecule-Perception Dataset

Due to the scarcity of large-scale mixture-odor perception correspondence data, our study innovatively employs single-molecule data to guide the odor prediction of mixtures. We curated a substantial dataset of single-molecule perception correspondences for model optimization, as detailed in Figure S1 (b).

The single-molecule odor perception dataset, derived from the Good Scents and Leffingwell databases[4-5], provides the interaction relationships between molecules and odors. These molecules are labeled with one or more odor descriptors. Both databases are publicly available and have been rigorously validated through biochemical experiments. Professional olfactory experts labeled the odorant molecules based on empirical testing results, offering detailed descriptions of the odor characteristics for each compound according to actual olfactory properties. Therefore, the obtained correspondence between molecules and odors is both validated and reliable.

We removed duplicate entries to ensure the consistency of each data point in the dataset. To minimize discrepancies and inconsistencies, we performed cross-validation across datasets. By retaining consistent and reliable data, we enhanced the overall accuracy and robustness of the final dataset. Furthermore, we excluded samples containing missing values and erroneous labels to avoid introducing biases, while also discarding entries with missing critical information to prevent noise interference. These procedures improved the overall data quality for subsequent analysis and model training. We ultimately compiled an accurate molecule-odor dataset, which includes 5,023 single molecules and 151 distinct odors. The dataset is provided as Supplementary Material 2.

**C. Mixture Similarity Rating Dataset**

To rigorously validate the accuracy of our proposed method in identifying mixture perception, we curated a mixture similarity dataset (as shown in Figure S1 (c)) sourced from existing literature [6-8]. This dataset provides detailed information on the chemical composition of the mixtures and the corresponding perceptual similarity ratings between mixture pairs. The dataset includes a total of 524 unique single molecules, forming 112 distinct mixtures, and contains 381 pairwise similarity comparisons. The similarity ratings for each mixture pair were obtained through multiple human perceptual trials, with data collected through stringent olfactory psychophysical experiments, ensuring the reliability and scientific validity of the data.

The complexity of the mixtures varies significantly, with the number of constituent molecules ranging from 4 to 43. The mixtures in the dataset span from simple combinations of a few molecules to complex mixtures consisting of up to several dozen molecules, reflecting the complexity and diversity of multi-molecule interactions in odor perception. In each experimental trial, subjects were presented with two mixture stimuli and asked to rate the perceptual similarity between them. The similarity ratings were based on the subjective judgment of the participants regarding the degree of similarity between the two odor mixtures, with scores ranging from 0 to 100, where 0 indicates completely different and 100 indicates identical.

Each subject underwent rigorous training, and every mixture pair in the dataset was independently rated multiple times to ensure the stability and reliability of the scores. This approach allowed us to accurately capture the perceptual similarity between different mixtures, providing a solid data foundation for the study of mixture odor perception. The dataset is provided as Supplementary Material 3. We used this dataset to evaluate the performance of our method, specifically to assess its ability to accurately predict the perceptual similarity between mixtures.

**D. Mixture–Perception Dataset**

To evaluate the performance of our method in real-world scenarios, specifically its ability to accurately identify the odor perception of multi-molecule mixtures, we curated a validation dataset derived from two authoritative sources: Essential Oil Safety database [9] and the Good Scents Company database. Essential Oil Safety database is widely recognized as an authoritative reference in the field of essential oils and fragrances, providing detailed chemical composition data based on extensive chromatographic analysis, covering the constituent molecules and respective concentration ratios in various essential oils. The Good Scents Company database is a standard resource in the flavor and fragrance industry, offering standardized odor perception for thousands of raw materials and natural mixtures. The mixture-perception dataset is shown in Figure S1 (d).

We extracted the chemical composition of natural mixtures from Essential Oil Safety database, recording the constituent molecules and corresponding concentration ratios. Subsequently, we retrieved the odor perception labels for these mixtures from the Good Scents Company database. This process enabled us to establish an accurate mapping between the multi-molecule mixtures and odor perceptions. Ultimately, we compiled a dataset of 172 real-world test samples, serving as an independent benchmark for validating our model's performance. This dataset is provided as Supplementary Material 4. By utilizing this dataset, we were able to effectively assess the model's performance in recognizing the odor perception of multi-molecule mixtures in real-world settings, to further validate the effectiveness and generalizability of our approach.

### E. Mixture-Response Curve Dataset

We manually curated experimentally measured neural response curves induced by multi-molecule mixtures from previously published studies. Specifically, we reviewed 22 studies [10-31] reporting olfactory receptor response experiments and extracted 35 mixture–response curve pairs. All data were derived from biochemical experiments and reflect dynamic biosignal responses generated by interactions between multi-molecule odor mixtures and olfactory receptors.

During data collection, we applied the following inclusion criteria. First, the stimulus had to be a chemically defined multi-molecule odor mixture rather than an undefined natural odor source. Second, the molecular components of each mixture and their concentration ratios had to be explicitly reported. Third, the study had to provide time-dependent neural response curves, namely the membrane potential of neurons. Fourth, the curve images or raw experimental data had to be of sufficient quality for subsequent digital extraction, thereby supporting external validation of the model. Ultimately, we assembled an external validation dataset containing 35 mixture–response curve pairs. Each mixture contained 3 to 7 single-molecule components, with distinct concentration ratios among different components. Thus, the dataset covered a certain degree of mixture complexity, including variations in molecular composition, component number, and concentration distribution. These characteristics enabled us to evaluate whether the proposed model could capture neural response patterns induced by realistic multi-molecule olfactory stimuli, which are jointly shaped by mixture composition and concentration.

Because different studies used different experimental platforms, stimulation protocols, signal recording methods, and normalization strategies, we standardized the collected curves before validation. For response curves presented only as images, we digitized the curve data. We then normalized the time axes and response amplitudes across datasets and unified the response direction according to the activation patterns reported in the original studies. Given the differences in experimental conditions across studies, we did not directly compare absolute signal amplitudes. Instead, we focused on the consistency between experimentally measured and model-predicted curves in terms of temporal morphology, dynamic trends, and response patterns. This strategy allowed us to more reasonably assess whether the model could reproduce biologically plausible response dynamics induced by multi-molecule mixtures under limited external validation data.

## II. Overall Workflow of Multi-Molecule Mixture Perception Recognition Approach

The objective of our study is to accurately identify the odor perception of multi-molecule mixtures, bridging the theoretical gap between chemical blending and perceptual formation. By integrating multi-level information encompassing molecule-receptor interactions, neural encoding, and perceptual semantics, we aim to provide a computational model for elucidating the generative mechanisms underlying olfactory perception. We innovatively proposed a deep learning framework that incorporates a bio-inspired olfactory mechanism to solve this problem. Due to the lack of sufficient mixture-perception datasets for direct model training, we guide the prediction of multi-molecule mixture perception by modeling the structural features of molecules and receptors, and utilizing the existing abundant single-molecule to perception data. The model is optimized by minimizing the discrepancy between the odor perception of the molecules and the response curves. Furthermore, we developed a fusion strategy for the mixture multi-receptor response curves. This strategy integrates concentration-dependent multi-molecule curves with attention-weighted multi-receptor curves, considering both the types and concentrations of molecules, as well as the types of receptors, to accurately establish the biological simulation of the response curves between the mixtures and multiple receptors. Finally, by comparing the consistency of single-molecule and multi-molecule response curves and dynamically assigning different weights to the odors, the odor perception of multi-molecule mixtures can be accurately recognized. This framework transcends the limitations of traditional black-box structure-to-perception mapping approaches. The overall steps of our method are outlined below. The formulation description is in the Section III.

**Step 1 (Figure 2a). Deep Learning-Based Prediction of Molecule-Receptor Response Curves.** Based on the molecule sequence features and the three-dimensional structure of olfactory receptors, we specifically designed deep learning models to capture their interaction patterns. Deep neural decision forests (DNDF) and hypergraph neural networks (HGNN) were employed to extract feature representations of molecules and receptors, respectively. The resulting feature vectors were subsequently fused and fed into the fully connected network to predict key characteristics of the neural response curves, thereby reflecting the interaction dynamics between each molecule–receptor pair. In addition, inspired by established olfactory biological mechanisms, we modeled the neural response curves using a decaying sine wave, which simulates the characteristic temporal firing dynamics of olfactory receptors following odorant stimulation. Collectively, this step provides the foundational basis for constructing mixture response curves and modeling the overall olfactory perception.

**Step 2 (Figure 2b). Aligning Discrepancies Between Response Curves and Odor Perceptions to Optimize the Model.** The differences in odor perception between molecules exhibit a strong consistency with the corresponding differences in their neural response curves: larger perceptual differences are typically associated with more pronounced discrepancies in response curves, and vice versa. Leveraging this intrinsic consistency, we jointly exploit single-molecule odor perception data and corresponding multi-receptor neural response curves by formulating the deviation between inter-molecule response curve differences and perception differences as the loss function for model optimization. By minimizing the discrepancy, the model can learn a stable and coherent mapping between the odor perception and neural response curve, thereby improving the prediction accuracy of molecule to receptor neural response curves and establishing a reliable foundation for multi-molecule mixture perception.

**Step 3 (Figure 2c). Attention-Weighted Multi-receptor Response Curve Fusion Strategy.** Olfactory perception arises from the coordinated activation of multiple olfactory receptors, with individual receptors contributing unequally to the final percept. Motivated by this biological principle, we incorporated an attention mechanism into the proposed framework to dynamically assign distinct weights to different olfactory receptors, thereby capturing relative importance in the perception formation process. Building upon this mechanism, we designed a multi-receptor response curve fusion strategy to effectively integrate the collective responses of multiple receptors. Through this strategy, we are able to construct single-molecule to multi-receptor response curves for model training in step 2, and further generate multi-molecule to multi-receptor response curves, which serve as critical inputs for subsequent modeling of mixture odor perception.

**Step 4 (Figure 2d). Concentration-Aware Mixture Response Curve Fusion Strategy.** Given that the constituent molecules in a mixture exhibit substantial variability in concentrations, and molecule concentration plays a critical role in shaping neural response curve profiles as well as the odor perception, we proposed a concentration-aware mixture response curve fusion strategy. Specifically, for molecules under different concentration conditions, we first constructed their individual response curves with each olfactory receptor based on Step1. We then employ a concentration-aware weighted curve fusion mechanism to capture the modulatory effects of concentration variations on receptor activation patterns. Ultimately, for the multi-molecule mixture, we can derive the specific interaction response curves between the mixture and each olfactory receptor. Subsequently, these receptor-level response curves are integrated using the attention-weighted multi-receptor response curve fusion strategy introduced in Step 3, yielding a unified response curve that effectively characterizes the overall odor perception of the multi-molecule mixture.

**Step 5 (Figure 2e). Response Curve Similarity Metrics Based Mixture Odor Perception Determination.** We leverage single-molecule odor perception as a prior to guide the identification of odor perception in multi-molecule mixtures. Specifically, by computing similarity metrics between the response curves of the mixture and those of individual single molecules, we adaptively assign weights to the odor perception vectors associated with different molecules, thereby reflecting differential contributions to the overall odor perception. These weighted odor vectors are subsequently aggregated, and a threshold selection is applied to determine the final perceptual outputs. Ultimately, the model is able to robustly predict the multi-label odor perception of the target mixture, achieving an accurate mapping from the neural response domain to the odor perception space and completing the final determination of mixture odor perception.

The specific descriptions of Step 1 to Step 5 are as follows:

## A. Deep Learning-based Prediction of Molecule-Receptor Response Curves

In this section, we employed the modified DNDF and HGNN to model molecules and olfactory receptors, respectively, extracting structure feature and designing fully connected neural networks to predict the neural response curves that characterize their interactions. Additionally, based on biological mechanisms, we used the decaying sine wave function to mathematically simulate the dynamic neural response characteristics elicited by molecule-receptor interactions. The specific methodology is outlined as follows:

### (1) Data Features of Molecule and Receptor

We utilized molecular sequence data and receptor structural data as features for both entities, and then employed a deep learning model for modeling. The data features of molecules and receptors are described as follows.

**Molecule Sequence Data.** The molecule sequence data is characterized as a hybrid feature set integrating physicochemical descriptors [32] and molecular fingerprints [33]. These components have vector dimensions of 1444 and 473, respectively, yielding a concatenated feature vector with a total dimensionality of 1917. This integration ensures a robust representation of both the physicochemical properties and topological structures of the molecules. Specifically, physicochemical descriptors characterize molecule properties that are essential for analyzing the relationship between molecule structure and biological activity. These include molecular weight, topological polar surface area, number of atoms, molecular refractivity, and others. These parameters facilitate the characterization of molecule behavior, interaction patterns, and biological activity. Furthermore, Molecule fingerprints are numerical representations that encode the structure of a molecule, capturing atom-based and bond-based features. These fingerprints represent structural information, such as functional groups, atom connectivity, and molecular substructures, which can be used for comparative analysis and similarity assessments between molecules.

**Receptor Structure Data.** To capture the 3D conformational information of olfactory receptors, each receptor was encoded using a comprehensive multi-attribute scheme that includes atomic coordinates, charges, masses, and bonding parameters, and other key physical features [34-35]. Data extraction was performed by synthesizing multiple structural file formats: residue and atom identities were derived from PDB files; spatial coordinates were obtained from GRO files; and force-field parameters (including charge, mass, bond/dihedral constraints) were extracted from TOP files. Collectively, these integrated features reflect both the inter-atomic physicochemical interactions and the spatial topological arrangement of residues, providing a rigorous biophysical basis for modeling the receptor's 3D structural features.

**(2) Deep Learning Modeling for Molecule and Receptor**

We developed deep learning models to model both molecules and receptors, aiming to accurately extract structural features. For molecule sequence data, we prioritized algorithms capable of effectively handling high-dimensional numerical feature vectors. Given the proven efficacy of Random Forest (RF)-based algorithms in processing such data, we adopted the Deep Neural Decision Forests (DNDF) model. DNDF effectively bridges the gap between the interpretability of RF-based methods and the representational learning capabilities of deep neural networks. By implementing stochastic and differentiable decision trees, DNDF allows for the global optimization of split node parameters via backpropagation. This mechanism enables DNDF to effectively capture complex nonlinear patterns within molecule sequence data.

For receptor structure data, to encode the complex 3D topology of olfactory receptors, we adopted a Hypergraph Neural Network (HGNN) framework. This architecture is particularly well-suited for capturing the spatial dependencies crucial for ligand binding. We represent the receptor structure as a hypergraph, defining atoms as nodes (attributed with properties such as type, charge, and mass) and chemical bonds as hyperedges (attributed with properties such as type, length, and dihedral angle). By employing a multi-layer graph convolution strategy, the model iteratively aggregates information from neighboring nodes, generating a representation that reflects the receptor's spatial conformation.

Therefore, DNDF and HGNN were selected as the backbone networks for molecule and receptor modeling, respectively, to accurately extract structure features. Additionally, we made specific structural modifications to these models to optimize their performance for the unique characteristics of the data. The detailed modifications of these architectures are provided in Section IV.

**(3) Single-Molecule to Single-Receptor Response Curves**

We designed a fully connected neural network that takes molecule and receptor features as input to predict the neural response curves (Figure S2). Additionally, we adopted a decaying sine wave function to mathematically simulate the dynamic neural response characteristics induced by single-molecule to single-receptor interactions.

The selection of this mathematical model is predicated on accurately recapitulating fundamental neurophysiological characteristics of the biological olfactory system. First, the model is designed to capture the mechanisms of neuronal adaptation and receptor desensitization. Physiological studies [36] demonstrated that under sustained odorant stimulation, olfactory receptor neurons do not maintain a constant firing rate. Instead, they typically exhibit a characteristic response pattern: a transient, high-intensity phasic response upon initial stimulation, followed by a rapid exponential decay toward baseline levels. This dynamic decay behavior reflects the receptor's adaptation process to prolonged stimulation. Consequently, we modeled this biophysical phenomenon using the exponential damping term within the wave function. Second, the model incorporates a periodic term to reflect the intrinsic temporal oscillations central to olfactory coding [37]. The olfactory system does not rely on static coding; rather, neural assemblies within the olfactory bulb and cortex generate rhythmic oscillatory activities. These oscillations function as an internal clock that synchronizes spike timing, thereby providing critical temporal windows for precise odor discrimination. The sine term in the wave function effectively characterizes these rhythmic fluctuations in neuronal excitability and their periodic variations along the time axis. Grounded in these biological mechanisms, we adopted a decaying sine wave function to mathematically simulate the dynamic neural response induced by single-molecule to single-receptor interactions, defined as follows: $R(t) = A \cdot e^{-\lambda t} \cdot \sin(\omega t + \varphi)$

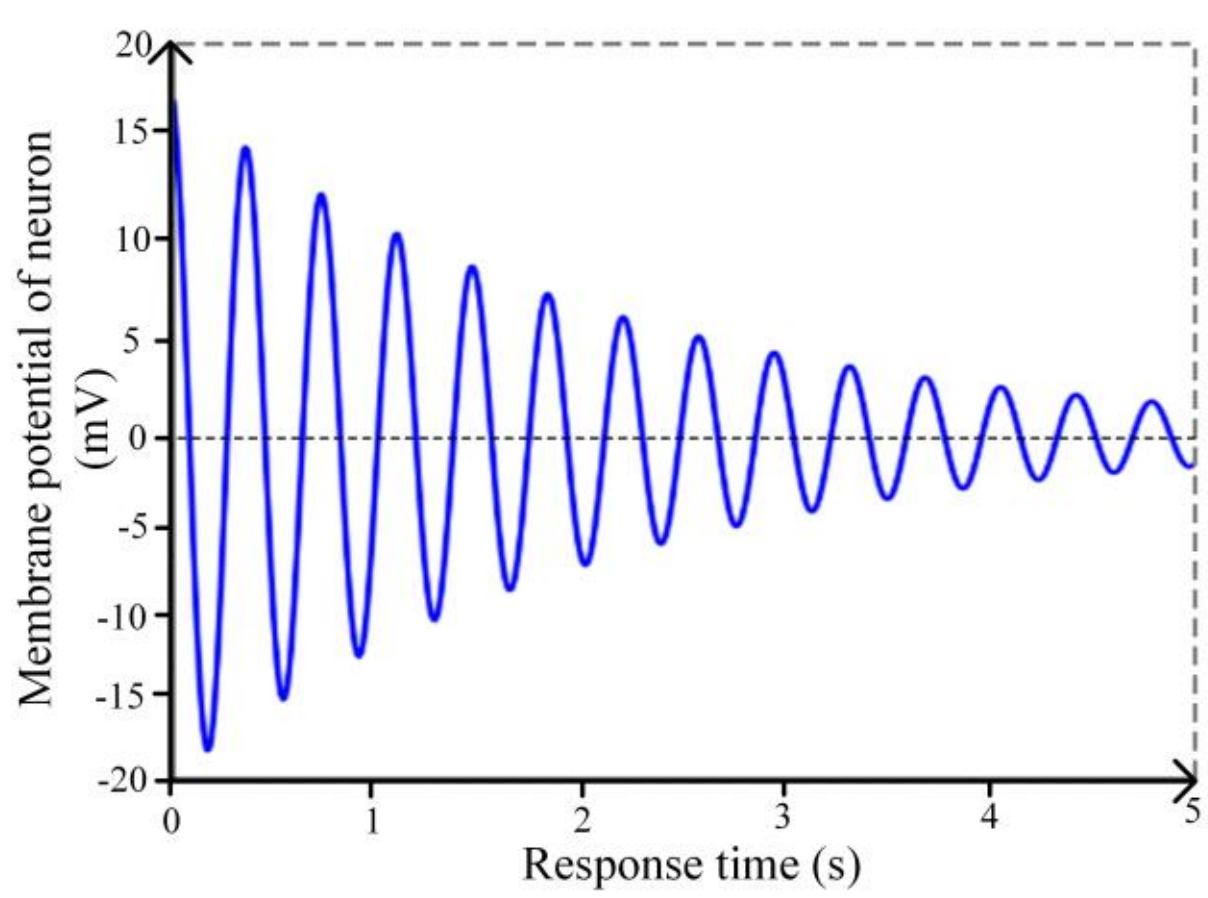

**Fig. S2. Decaying Sine Wave Response Curve**

The decaying sine wave curve. The horizontal axis represents the response time, and the vertical axis represents the membrane potential of the olfactory receptor neurons. The morphology of this curve is determined by four key parameters: Amplitude (A): Determines the initial peak intensity of neural activation. Damping Coefficient (λ): Governs the rate of exponential decay towards baseline, simulating the receptor desensitization process. Angular Frequency (ω): Characterizes the rhythm of oscillation, reflecting the firing frequency of the neural assembly. Phase Shift (ϕ): Defines the initial state of signal response delay along the time axis. Considering the biological processes involved in molecule-receptor interactions [38], we set the range of membrane potential to ±20mV with the maximum response time limited to within 5 seconds.

**B. Align Discrepancy Between Response Curves and Odor Perceptions to Optimize Model**

Based on the aforementioned steps, we can predict the neural response curves for single molecule and single olfactory receptor using the deep learning model. Moreover, by applying the method described in Section II-C, the fused response curve of a single molecule across multiple receptors—representing the final molecular response curve—can be derived. However, a critical challenge remains: the lack of abundant real-world neural response recordings to serve as direct supervisory labels for training and optimizing the model. Therefore, we innovatively proposed a perception-guided contrastive learning-based unsupervised training strategy. This approach guides the generation of molecule-receptor response curves by aligning the discrepancies between the neural response curves of molecules and corresponding odor perceptions. As shown in Figure S3, the details are as follows:

Our objective is to align the discrepancies between neural response curves and odor perception differences. Specifically, a stringent mapping exists between the perceptual attributes of odorant molecules and the neural dynamical patterns they evoke. When the odor perceptions of two molecules differ substantially, the corresponding neural response curves likewise exhibit pronounced morphological differences, and vice versa [39]. This implies that the divergence in neural response trajectories is not merely a feedback consequence of distinct chemical stimuli, but rather constitutes a biological substrate through which the brain differentiates odor percepts. Therefore, we used the set of single-molecule odor labels as a discriminative indicator to align and minimize the deviation between response curves and odor perception. For each pair of molecules, we calculate the differences between response curves using Mean Absolute Error (MAE) and measure odor perception differences using cosine distance. These two metrics are aligned using the Pearson correlation coefficient (Figure S3, top). The deviation between the molecular response curve differences and the perceptual differences serves as the loss function, and the deep learning model is optimized by minimizing this loss (Figure S3, bottom). The dataset contains 5,023 single molecules, and we computed the pairwise deviation for all molecule pairs, yielding approximately twelve million deviations, which are sufficient for model optimization.

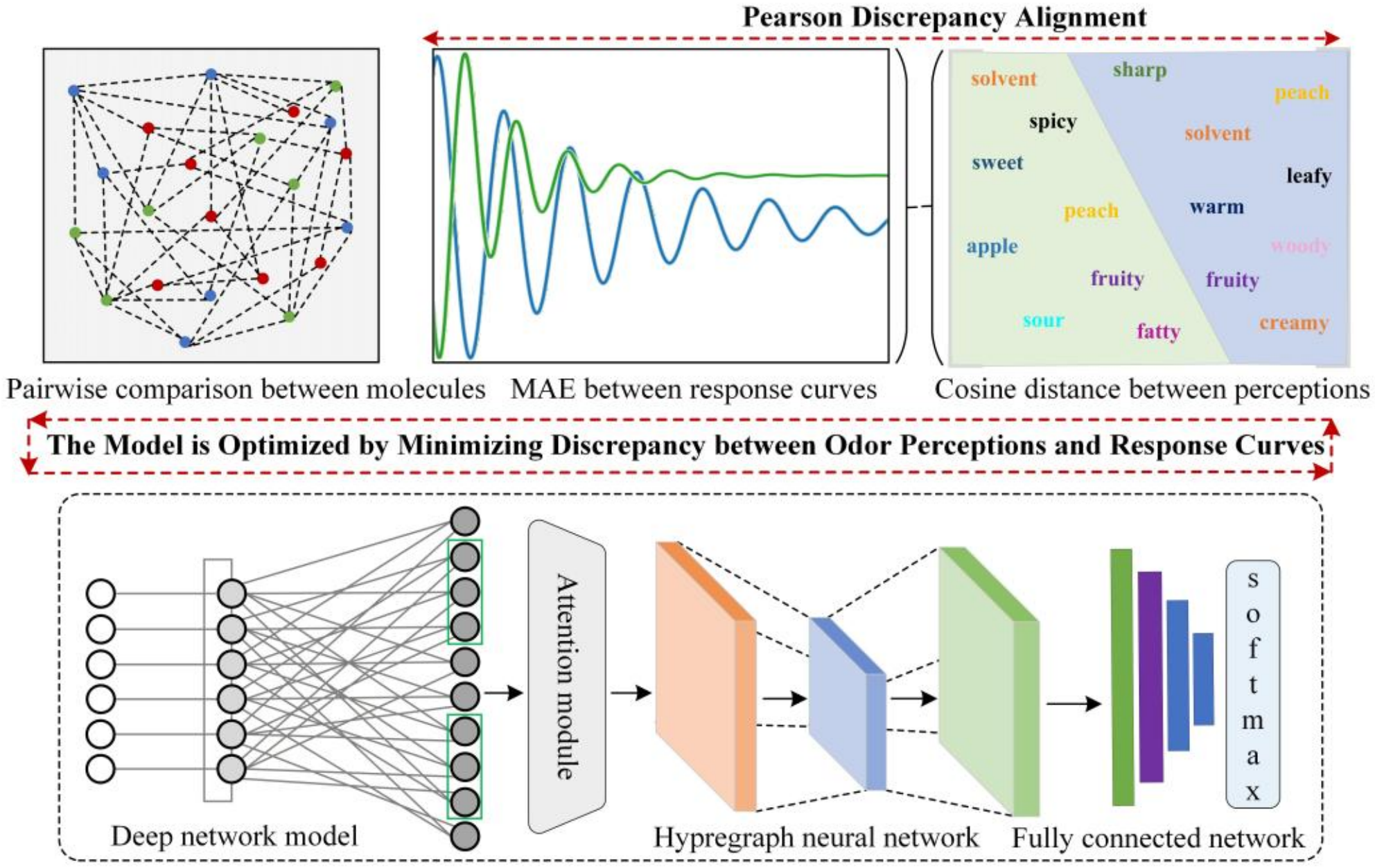

**Fig. S3. Align the Discrepancy to Optimize the Model.** We optimized the model by aligning the discrepancies between the neural response curves and odor perceptions for each pair of molecules, using this alignment deviation as the loss function to guide the generation of molecule-receptor response curves. The ultimate optimization objective is to ensure that the difference in neural response curves between any two molecules is isomorphic to their perceptual odor difference. In other words, a large odor perception difference corresponds to a large difference in neural response curves, and vice versa.

### C. Attention-Weighted Multi-Receptor Response Curve Fusion Strategy

To accurately capture the differential contributions of distinct receptors to the odor perception of specific molecules, an attention module was integrated into the deep learning model, as the influence of each receptor on final odor perception varies [40]. For a given odorant molecule, not all receptors contribute equally to the formation of the percept; rather, a small subset of key receptors dictates the final olfactory identity. Consequently, the attention module is designed to emulate this biological selectivity mechanism. By adaptively learning and assigning weights based on molecule-receptor interaction features, the module effectively amplifies the salience of key receptors while attenuating interference from redundant receptors. Specifically, we developed the attention module, which concatenates the molecule and receptor features as input and predicts a relevance score via a non-linear mapping. Subsequently, to ensure that the weights satisfy the properties of a probability distribution, we employed the softmax function to normalize these scores, ensuring that the sum across all receptor sets equals 1, thereby yielding the final attention weights. The attention module is integrated into the main framework for joint training and optimization.

As shown in Figure S4, since olfactory perception relies on the collective activity of the entire receptor repertoire rather than the isolated signals from individual receptors, it is essential to integrate the neural responses generated by all receptors. For the response curves of multiple receptors, we used the attention-weighted multi-receptor response curve fusion strategy. We simulated the biological signal integration process by superimposing the individual response curves across the temporal domain, thereby preserving the dynamic temporal characteristics of neural activity [41]. Additionally, we developed a time-point weighted linear superposition approach. For the response curves elicited by a molecule across different receptor repertoire, we performed a weighted average of the individual response curves according to corresponding attention weights, calculating the response value at each time point, ultimately generating the global response curve for all receptors.

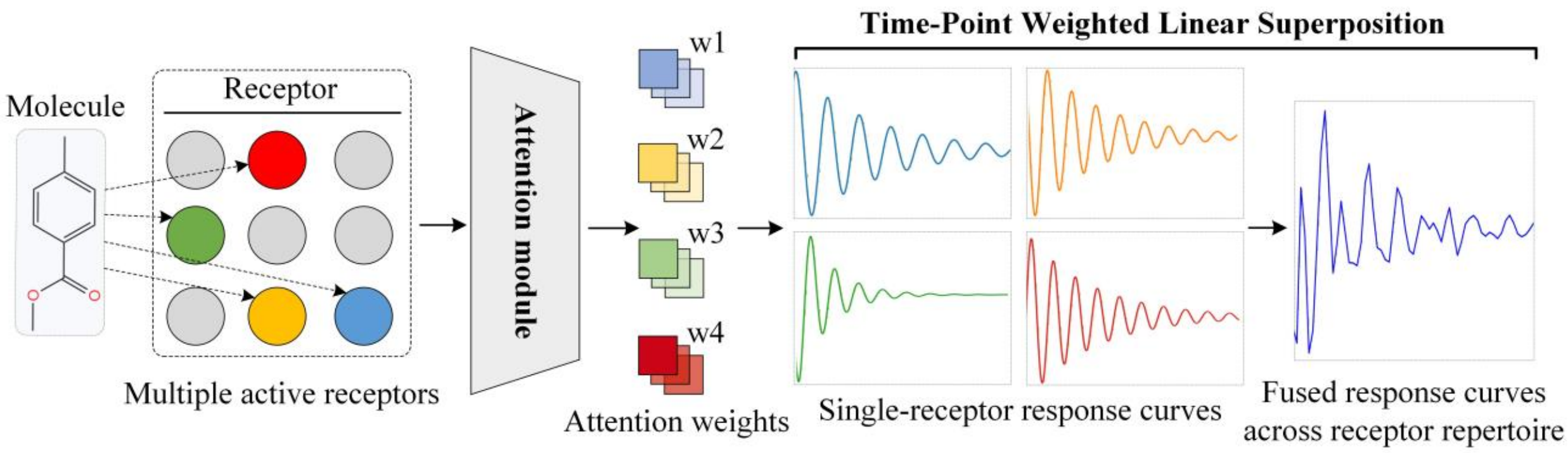

**Fig. S4. Attention-Weighted Response Curve Fusion Strategy.** Due to the unequal contributions of different olfactory receptors to odor perception, a simple unweighted average of the response curves would dilute key response features. Therefore, we introduced the attention-based weighting mechanism to enhance the contribution of key receptors. The attention module dynamically assigns different weights to each active receptor and performs a weighted average of the individual response curves based on corresponding weights, generating the global response curve for the entire receptors.

### D. Concentration-Aware Mixture Response Curve Fusion Strategy

Using the designed deep learning model above, we can accurately predict the neural response curves elicited by single molecule with multiple receptors. In this section, our goal is to generate the fused response curves for the entire olfactory receptor repertoire triggered by a multi-molecule mixture. As shown in Figure S5, this process is divided into two stages: Stage 1. Multi-molecule to single-receptor interaction: This involves the interaction of multiple molecules with a single olfactory receptor, generating the receptor's response curve to the mixture. Stage 2. Multi-receptor integration: The response curves generated by all receptors are globally integrated to yield the collective neural response profile triggered by the mixture, using the proposed attention-weighted response curve fusion strategy. Stage 2 has been introduced in Section III-B, thus the focus here will be on the multi-molecule to single-receptor interaction process.

In complex mixture environments, a single olfactory receptor will be influenced by the competitive binding of multiple molecules. Therefore, this step aims to analyze the synergistic effects of multiple molecules, particularly by constructing the fused neural response curve induced by the mixture at the single receptor level [42]. Considering the significant impact of molecule types and concentrations in the mixture on odor perception, we proposed the concentration-based mixture response curve fusion strategy. As shown in Figure S5, firstly, we calculated the concentration proportion of each molecule in the mixture, normalize it (a). Next, using the deep learning models from Sections III-B and C, we predicted the neural response curves for individual molecules and single olfactory receptors. We then used concentration as the weight for the molecule's response curve to measure its competitive binding advantage at the receptor binding site (b). Finally, the weighted response curves of all molecules are fused through temporal linear superposition. At each time point, the response values of all molecules are summed and averaged according to their concentration weight, thereby generating the global response curve for the single receptor in response to the multi-molecule mixture (c).

After obtaining the response curves at the single-receptor level, we can integrate them using the attention-weighted multi-receptor response curve fusion strategy outlined in Section III-B. Through this process, we can obtain the global neural response curve for the multi-molecule mixture across the entire receptor repertoire.

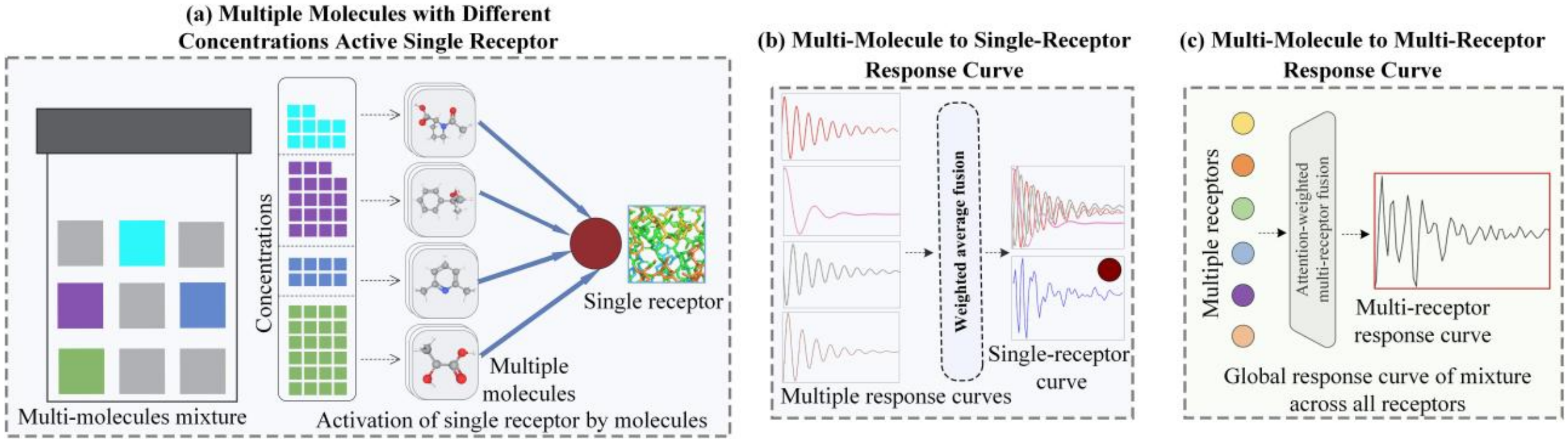

**Fig. S5. Multi-Molecule to Multi-Receptor Fusion.** We first calculated the concentration proportions of each molecule in the mixture and performed normalization (a). Then, using a weighted averaging method, we fused the response curves of individual molecules interacting with the receptor. We obtained the multi-molecule to single-receptor response curve (b). Finally, we can obtain the global neural response curve for the multi-molecule mixture across the entire receptor repertoire based on attention-weighted multi-receptor response curve fusion strategy (c).

### E. Response Curve Similarity Metrics Based Mixture Odor Perception Determination

After generating the global neural response curve of the multi-molecule mixture, the final task is to decode the corresponding perceptual odor from this response profile. We adopt a similarity-based inference strategy grounded in the response curve to odor perception isomorphism truth, which posits that molecule stimuli eliciting similar neural response patterns tend to share similar odor perceptions [43]. Consequently, we utilized a well-annotated dataset, which includes single-molecules and corresponding odor perceptions, as a reference repository. By measuring the similarity between the neural responses of the mixture and those of single molecules, we transfer odor perception from the single-molecule domain to the mixture domain.

The specific inference procedure is as follows: 1) Similarity Metric Calculation: First, we calculated the geometric similarity between the mixture response curve and each single-molecule response curve using mean absolute error (MAE), quantifying similarity scores. 2) Weight Normalization: The resulting similarity scores are normalized and used as contribution weights for the odor perception of each single molecule. A higher similarity indicates that the odor perception of the corresponding single molecule is more similar to that of the mixture. 3) Weighted Semantic Aggregation: Using the computed weights, we aggregated the odor perceptions of all single molecules via weighted averaging, yielding the odor probability distribution in the perceptual space for the mixture. 4) Thresholding: Finally, a confidence threshold is applied for filtering. Odor labels whose aggregated probabilities exceed this threshold are retained, collectively forming the perceptual odor set for the multi-molecule mixture.

## III. Formulation Implementation

The objective of our study is to accurately identify the odor perception corresponding to multi-molecule mixtures. To achieve this, we first design a deep learning model to characterize the structural features of molecules and olfactory receptors, and simulate the neural response curves elicited by their interactions based on biologically-inspired mechanisms. To address the lack of extensive real-world paired data between multi-molecule mixtures and odor perceptions, we propose an unsupervised training strategy that leverages existing single-molecule data by aligning discrepancies between response curves and odor perceptions to optimize the model. Furthermore, considering the differential contributions of distinct olfactory receptors to odor perception, we design an attention-weighted multi-receptor response curve fusion strategy, which dynamically assigns weights to individual receptors to achieve effective integration of response curves at the multi-receptor level. In addition, given the significant influence of molecule composition and concentration ratios within mixtures on the final odor perception, we develop a concentration-aware mixture response curve fusion strategy. Combined with the receptor-level fusion approach, this enables the accurate construction of global response curves between multi-molecule mixtures and the olfactory receptor repertoire. Finally, based on response curve similarity metrics, we utilize single-molecule odor perceptions as prior knowledge to guide the identification of odor perceptions in multi-molecule mixtures.

The following sections elaborate on the mathematical formulations for each step of the proposed methodology, corresponding to Steps 1 through 5 in the main text.

### A. Deep Learning-based Prediction of Molecule-Receptor Response Curves

We designed a deep learning model to characterize both molecules and receptors, extracting structural features. In accordance with biological mechanisms, a decaying sine wave model was employed to simulate the neural response curves resulting from molecule-receptor interactions.

#### (1) Structure Representation of Molecule and Receptor

Firstly, each molecule $M$ is represented by a high-dimensional feature vector $\mathbf{f}_M \in \mathbb{R}^{1917}$, molecule sequence data, which is concatenated from two categories of data:

$$\mathbf{f}_M = [\mathbf{p}][\mathbf{q}]$$

$\mathbf{p}$ represents the physicochemical descriptor data, which includes 1,444 features such as molecular weight, topological polar surface area, number of atoms, and molecular refractivity. These features are used to describe the physicochemical properties and biological activity of the molecule.

$\mathbf{q}$ represents the molecular fingerprint data, which maps the molecular structure to a binary vector representing its atom types, chemical bonding patterns, functional groups, and molecular substructures. This feature vector integrates the physical and chemical properties with the topological structure of the molecule, providing rich input for subsequent deep learning modeling.

Secondly, to accurately capture the three-dimensional conformational characteristics of the olfactory receptor, we represent the receptor structure using a graph structure. The receptor $R$is represented as $\boldsymbol{G} = (\boldsymbol{V}, \boldsymbol{E})$, where:

Node set $\boldsymbol{V}$**:** Each atom $\boldsymbol{v_i}$ corresponds to a node, with its attribute vector $\mathbf{a}_i \in \mathbb{R}^{\boldsymbol{d_a}}$ containing properties such as atom type (e.g., C, N, O), partial charge, atomic mass, van der Waals radius, etc.

Hyperedge set $\boldsymbol{E}$**:** Each chemical bond (including covalent bonds, hydrogen bonds, etc.) is represented as a hyperedge $\boldsymbol{e_j}$, with its attribute vector $\mathbf{b}_j \in \mathbb{R}^{\boldsymbol{d_b}}$containing geometric and energy parameters such as bond type (single bond, double bond, etc.), bond length, bond angle, and dihedral angle.

**(2) Feature Extraction and Modeling**

Considering that molecular sequence data consist of high-dimensional numerical feature vectors, we employed an enhanced deep neural decision forest (DNDF) to model the structural features of molecules. DNDF integrates differentiable decision trees with the traditional random forest model, combining the discriminative power of random forests with the representation learning capabilities of deep neural networks. During the forward propagation process, each sample $\mathbf{f}_M$ is routed through multiple decision trees, and the final output is the weighted sum of embeddings at all leaf nodes:

$$\mathbf{F}_M = \sum_{t=1}^{T} \pi_t(\mathbf{f}_M) \cdot l_t$$

where $\pi_t(\mathbf{f}_M)$ represents the probability that the sample belongs to the $t$-th leaf node, and $l_t$ is the embedding vector of the leaf node. The splitting function at each tree node is modeled using a differentiable sigmoid function:

$$\theta_t = \sigma(\boldsymbol{w}_t^T \mathrm{f}_M + b_t)$$

All parameters $\theta_t$ of the tree nodes are globally optimized using backpropagation and gradient descent, enabling the model to adaptively capture complex nonlinear patterns within the molecular features.

Subsequently, the structural features of receptors are modeled using an enhanced hypergraph neural network (HGNN). The core operation of HGNN is hypergraph convolution, with the update rule for node representation at the $l + 1$ layer defined as:

$$\mathbf{H}^{(l+1)} = \sigma(\mathbf{D}_v^{-1}\mathbf{HWBD}_e^{-1}\mathbf{H}^{(l)}\mathbf{\Theta}^{(l)})$$

$\mathbf{H}^{(l+1)} \in \mathbb{R}^{|\mathcal{V}|\times d_l}$ denotes the feature matrix of all nodes at layer $l$. $\mathbf{D}_v \in \mathbb{R}^{|\mathcal{V}|\times|\mathcal{V}|}$ and $\mathbf{D}_e \in \mathbb{R}^{|\mathcal{E}|\times|\mathcal{E}|}$ represent the node degree matrix and hyperedge degree matrix, respectively, which are used for normalization. $\mathbf{B} \in \mathbb{R}^{|\mathcal{V}|\times|\mathcal{E}|}$ is the incidence matrix of the hypergraph, where $\mathbf{B}_{ij} = 1$ indicates that node $i$ belongs to hyperedge $j$. $\mathbf{W}$ is a learnable hyperedge weight matrix. $\mathbf{\Theta}^{(l)} \in \mathbb{R}^{d_l \times d_{l+1}}$ is a trainable weight matrix for layer $l$. $\sigma$ denotes a nonlinear activation function.

Following $L$ convolutional layers, a global average pooling operation is employed to obtain a comprehensive representation of the receptor:

$$\mathbf{F}_R = \frac{1}{|\mathcal{V}|}\sum_{i=1}^{|v|} \mathbf{H}_i^{(L)}$$

**(3) Generation of Neural Response Curves**

The molecule features $\mathbf{F}_M$ and receptor features $\mathbf{F}_R$ are concatenated and passed through a fully connected neural network (FCN) to predict the parameters of the decaying sine wave:

$$\mathbf{z} = FCN([\mathbf{F}_M \| \mathbf{F}_R; \Theta_{FCN}]), \;\; \mathbf{z} \in \mathbb{R}^4$$

where $\mathbf{z} = [A, \lambda, \omega, \varphi]^\top$ represents the amplitude, damping coefficient, angular frequency, and phase shift, respectively. The FCN consists of multiple hidden layers utilizing ReLU activation functions, with dropout regularization applied to mitigate overfitting.

Consistent with established findings in olfactory neurophysiology—where neurons demonstrate a characteristic pattern of rapid activation, oscillatory firing, and progressive decay in response to odorant stimuli—we employ a decaying sinusoidal function to model the neural response curve:

$$C(t) = A \cdot e^{-\lambda t} \cdot \sin(\omega t + \varphi), \;\; t \geq 0$$

Each parameter in this formulation possesses a distinct biological interpretation: **Amplitude** $A$: Reflects the peak firing intensity during initial activation, correlating with ligand–receptor binding affinity. **Damping coefficient** $\lambda$: Governs the rate of exponential decay toward the baseline, simulating the process of receptor desensitization. **Angular frequency** $\omega$: Determines the oscillation rhythm, corresponding to the synchronized firing frequency of neuronal assemblies in the olfactory cortex. **Phase shift** $\varphi$: Represents the temporal delay in response onset, capturing the initial state of signal transduction and neuronal activation.

To ensure physiological plausibility, the following constraints are applied to the response function based on experimental observations of olfactory receptor neurons:

$$A \in [0, 50]\ \text{Hz}$$

$$\lambda \in [0.1, 5]\ \text{s}^{-1}$$

$$\omega \in [\pi, 15\pi]\ \text{rad/s}$$

$$t \in [0, 10]\ \text{s}$$

These bounds confine the predicted curves within biologically feasible ranges, thereby improving the model's reliability in simulating realistic neural dynamics.

**B. Aligning Discrepancies Between Response Curves and Odor Perception to Optimize the Model**

Due to the limited availability of paired data linking neural response curves and odor perceptions for model training, we adopt an unsupervised contrastive learning approach in this phase. By utilizing existing odor perception labels of individual molecules, we guide the generation of response curves, thereby refining the neural network model developed in Step 1. Based on the model introduced in Step 1 and the attention-weighted multi-receptor response curve fusion strategy (elaborated in Step 3), the integrated neural response curve $c_i$ for each molecule $M_i$ interacting with a set of olfactory receptors $\mathcal{R}$ is predicted as follows:

$$c_i = \sum_{j=1}^{|R|} \alpha_{i,j} \cdot C_{i,j}(t)$$

Here $C_{i,j}(t) = A_{i,j} \cdot e^{-\lambda_{i,j}t} \cdot \sin\left(\omega_{i,j}t + \varphi_{i,j}\right)$, $\alpha_{i,j}$ represents the attention weight assigned to receptor $j$ for molecule $i$, reflecting the receptor's relative contribution to the overall perceptual response.

The objective is to align the generated response curves with perceptual differences despite the absence of ground-truth neural response labels. Specifically, the model is optimized such that the discrepancy between predicted response curves correlates with differences in odor perception profiles across molecules. This perceptual-guided approach ensures that molecules with distinct odor perceptions yield divergent response curves, while perceptually similar molecules produce analogous neural response patterns.

**(1) Quantifying Discrepancy Between Response Curves and Odor Perceptions**

For any two molecules $M_a$ and $M_b$, the difference between their neural response curves $C_a(t)$ and $C_b(t)$ is quantified using the Mean Absolute Error (MAE) over the observation period $T$:

$$D_C(a,b) = \frac{1}{|T|}\int_0^T |C_a(t) - C_b(t)| d_t$$

The perception dissimilarity is measured via the cosine distance between perceptual label vectors $\mathbf{P}_a$ and $\mathbf{P}_b$:

$$D_P(a,b) = 1 - \frac{\mathbf{P}_a \cdot \mathbf{P}_b}{\|\mathbf{P}_a\|\|\mathbf{P}_b\|}$$

To align the response curve differences with perceptual differences, we compute the pearson correlation coefficient $r$ across all molecule pairs$(a,b)$:

$$r = \frac{\sum_{a<b}(D_C(a,b) - \overline{D}_C)(D_P(a,b) - \overline{D}_P)}{\sqrt{\sum_{a<b}(D_C(a,b) - \overline{D}_C)^2 \sum_{a<b}(D_P(a,b) - \overline{D}_P)^2}}$$

where $\bar{D}_C$ and $\bar{D}_P$ denote the mean values of $D_C$ and $D_P$, respectively, over all unique molecular pairs.

The objective is to maximize $r$, thereby enforcing a strong positive correlation between neural response divergence and perceptual dissimilarity. Accordingly, the alignment loss is defined as the negative correlation coefficient:

$$\mathcal{L}_{align} = -r$$

Minimizing this loss encourages the model to generate response curves whose pairwise differences are consistent with the corresponding differences in odor perception.

**(2) Model Optimization via Contrastive Learning**

To enhance the model's ability to discriminate between perceptually similar molecules, a contrastive learning component is introduced. For each anchor molecule $M_a$, its positive sample $M^+$ is defined as the molecule with the most similar odor perception profile, while the negative sample $M^-$ corresponds to the molecule with the most dissimilar perception. The contrastive loss is formulated as:

$$\mathcal{L}_{contrast} = \max(0, D_C(M_a, M^+) - D_C(M_a, M^-))$$

The total loss function is then defined as a weighted combination of the alignment loss and the contrastive loss:

$$\mathcal{L} = \mathcal{L}_{align} + \mathcal{L}_{contrast}$$

The parameters $\Theta$ of the deep learning model are optimized via backpropagation by minimizing the total loss:

$$\Theta^* = arg \min_{\Theta} \mathcal{L}$$

The dataset comprises $N = 5{,}023$ molecules, yielding a total of

$$\binom{N}{2} = \frac{N(N-1)}{2} \approx 12.6 \times 10^6$$

unique molecule pairs. This corresponds to approximately 12.6 million difference pairs, providing a robust and comprehensive supervisory signal for model optimization even in the absence of explicit neural response labels.

### C. Attention-Weighted Multi-Receptor Response Curve Fusion Strategy

#### (1) Attention Module for Receptor Contribution Scoring

To adaptively quantify the differential contribution of each olfactory receptor to odor perception, an attention mechanism is integrated into the deep learning framework. Let: $\mathbf{F}_M \in \mathbb{R}^{d_M}$ denote the feature vector of the input molecule, $\mathbf{F}_R \in \mathbb{R}^{d_R}$ denote the feature vector of the olfactory receptor.

The molecule-receptor pair is first concatenated and projected through a non-linear transformation to compute a raw relevance score:

$$s_j = \mathbf{w}^T \cdot \sigma(\mathbf{W} \cdot [\mathbf{F}_M \| \mathbf{F}_R] + \mathbf{b})$$

where $W \in \mathbb{R}^{d_a \times (d_M + d_R)}$ is a learnable weight matrix. $b \in \mathbb{R}^{d_a}$ is a bias vector. $\sigma(\cdot)$ denotes the ReLU activation function. $\mathbf{w} \in \mathbb{R}^{d_a}$ is the output projection vector. $[\cdot \| \cdot]$ indicates vector concatenation.

The raw scores are then normalized across all receptors using softmax function to obtain the final attention weights:

$$\alpha_j = \frac{\exp(s_j)}{\sum_{k=1}^{|R|} \exp(s_k)}$$

These weights satisfy $\alpha_j \in [0,1]$and $\sum_{j=1}^{|\mathcal{R}|} \alpha_j = 1$, representing a probability distribution over the receptor set.

#### (2) Weighted Fusion of Multi-Receptor Response Curves

For a given molecule $M$, the individual neural response curve elicited by receptor $R_j$ is defined as:

$$C_j(t) = A_j \cdot e^{-\lambda_j t} \cdot \sin(\omega_j t + \varphi_j)$$

where $A_j, \lambda_j, \omega_j, \varphi_j$ are predicted parameters from Step 1.

The global response curve $c_{(t)}$, integrating contributions from all receptors, is computed via a time-point-wise attention-weighted superposition:

$$c_{(t)} = \sum_{j=1}^{|R|} \alpha_j \cdot R_j(t)$$

This formulation ensures that receptors with higher attention weights exert greater influence on the fused response, while preserving the temporal dynamics of each individual curve.

**D. Concentration-Aware Mixture Response Curve Fusion Strategy**

**(1) Concentration-Weighted Fusion at Single-Receptor Level**

The concentration of each component molecule in a mixture significantly influences odor perception. To incorporate this effect into neural response modeling, we introduce concentration-proportional weighting into the response curve generation process. Consider a multi-molecule mixture M = {$M_1$, $M_2$ ,…, $M_m$} comprising $m$ distinct odorant molecules, each present at a concentration $c_i > 0$. The total concentration of the mixture is:

$$\epsilon_{total} = \sum_{i=1}^{M} \epsilon_i$$

The normalized concentration proportion for molecule $M_i$ is given by:

$\beta_i = \frac{\epsilon_i}{\epsilon_{total}}$, such that $\sum_{i=1}^{M} \epsilon_i$ =1

Here, $\beta_i$ represents the concentration-based weight assigned to molecule $M_i$, reflecting its relative abundance and, consequently, its competitive binding advantage at the receptor binding site under multi-ligand conditions.

For each molecule $M_i \in \mathcal{M}$, the single-molecule to single-receptor response curve $C_{i,j}(t)$ for receptor $R_j$ is predicted using the model described in Sections B and C: Subsequently, the response curves are weighted according to respective concentration proportions. The mixture response curve for receptor $C_j$, denoted $\tilde{C}_j(t)$, is obtained via temporal linear superposition of the individual response curves, weighted by $w_i$:

$$\tilde{C}_j(t) = \beta_i \cdot C_{i,j}(t)$$

Thus, the concentration-modulated neural response curve of a mixture at the level of a single receptor is generated.

**(2) Integration Across the Olfactory Receptor Repertoire**

After obtaining the mixture response curve $\tilde{C}_j(t)$ for each receptor $R_j$, we apply the attention-weighted multi-receptor fusion strategy outlined in Step 2 to integrate responses across the entire olfactory receptor repertoire. The global mixture response curve $C_{\text{mix}}(t)$ is computed as:

$$C_{mix}(t) = \sum_{j=i}^{N} \alpha_j \cdot \tilde{C}_j(t)$$

where $\alpha_j$ denotes the attention weight for receptor $R_j$, learned by the attention module described in Step 2. This weight represents the perceptual relevance of receptor $R_j$ for the given mixture.

Through this two-stage fusion process, we obtain the comprehensive neural response curve elicited by a multi-molecule mixture across the entire receptor repertoire, effectively simulating the integrated olfactory signal that underlies complex odor perception.

**E. Mixture Odor Perception Determination Based on Response Curve Similarity Metrics**

We determine the odor perception of a mixture by comparing the similarity of its neural response curves with those of well-annotated single molecules. Let $C_{\text{mix}}$ denote the global neural response curve of a target mixture, generated via the concentration-aware fusion strategy and attention-weighted multi-receptor response curve fusion strategy. The well-annotated reference dataset of single molecules:

$$D_{ref} = \{(C_k, \mathbf{p}_k)\}_{k=1}^{K}$$

where $C_k(t)$ is the neural response curve of the $k$-th reference molecule, which is accurately predicted by the aforementioned model. $\mathbf{p}_k \in \mathbb{R}^L$is its corresponding odor perception vector, with each dimension representing the intensity of a perceptual attribute (e.g., floral, fruity, woody, etc.).

Subsequently, we compute the cosine similarity $S_k$ between the mixture response curve $C_{\text{mix}}$ and each reference curve $C_k$. To convert the similarity scores into interpretable contribution weights, we apply a softmax normalization:

$$w_k = \frac{\exp(\gamma \cdot S_k)}{\sum_{j=1}^{K} \exp(\gamma \cdot S_j)}$$

where $\gamma > 0$ is a temperature parameter controlling the sharpness of the weight distribution. The weights satisfy:

$$w_k \in [0,1], \sum_{k=1}^{K} w_k = 1$$

A higher value of $w_k$ indicates that the odor perception of reference molecule $k$ contributes more significantly to the perceptual profile of the mixture. The predicted odor perception vector for the mixture, $\mathbf{p}_{mix}$, is obtained via weighted linear combination of the reference perception vectors:

$$\mathbf{p}_{mix} = \sum_{k=1}^{K} w_k \cdot \mathbf{p}_k$$

Thus, $\mathbf{p}_{\text{mix}} \in \mathbb{R}^L$ represents a probability-like distribution over perceptual attributes.

To extract a discrete set of dominant odor labels, a confidence threshold $\theta \in [0,1]$is applied:

$$\mathcal{O}_{mix} = \left(\ell \middle| p_{mix}^{\ell} \geq \theta\right)$$

where $p_{\text{mix}}^{(\ell)}$ denotes the $\ell$-th element of $\mathbf{p}_{\text{mix}}$, corresponding to perceptual attribute $\ell$. The threshold $\theta$ can be adjusted based on desired specificity and recall.

## IV. Overall Network Model

### A. Improvements to DNDF

As illustrated in Figure S6 (a), Given the varying importance of features in molecular sequence data and inherently high sparsity, we replaced the CNN module in the original DNDF with a feature extraction module comprising a multi-head self-attention mechanism and a fully connected network (FCN). This module treats molecular sequences as high-dimensional sparse numerical inputs and effectively extracts informative feature vectors.

**1) Multi-Head Self-Attention Mechanism.** We incorporate a multi-head self-attention mechanism to capture diverse and critical features embedded in molecule sequence, such as hydrophobicity, steric effects, and electronic properties, while dynamically assigning importance weights to each descriptor. Molecule sequence (e.g., MW, Pol, Zagreb index, and MACCS fingerprints) originates from heterogeneous dimensions and exhibits complex interdependencies. The self-attention mechanism automatically learns these correlations and generates context-aware representations, enabling the model to build meaningful feature embeddings. Unlike traditional neural networks, which treat all input features equally, the attention mechanism allows the model to focus more on informative features while attenuating the influence of irrelevant or noisy ones. This is particularly beneficial for molecular modeling tasks, where the accurate identification of key features—such as aromaticity and polarity—is crucial. The multiple attention heads enhance the model's representational capacity by jointly attending to various subspaces of physicochemical information, such as logP and molar refractivity. Furthermore, by dynamically modulating the importance of each descriptor based on the molecular structure, the model is able to capture non-linear and high-order interactions among molecule sequence, thereby significantly improving the predictive performance for molecular property estimation.

**2) Fully Connected Network (FCN).** To efficiently extract and transform high-dimensional molecular sequence features—with input dimensions reaching up to 2,000—we employ a multi-layer fully connected network (FCN) architecture. The FCN transforms sparse, high-dimensional fingerprint data (e.g., MACCSFP150, SubFP304) into compact and dense feature representations. Each layer applies non-linear transformations using ReLU activation functions to highlight informative features and suppress irrelevant noise, effectively identifying fingerprint positions associated with specific pharmacophores or structural motifs. Compared to directly using raw sparse input features, this approach significantly enhances training efficiency and accelerates convergence. Furthermore, when the attention mechanism highlights key descriptors such as molecular weight (MW) and polarity, the FCN further captures their high-order interactions with other latent or weak features, thereby improving the model's predictive performance. Given the typically high dimensionality of molecule sequence—comprising hundreds of dimensions—the FCN plays a critical role in dimensionality reduction, producing a unified, compact, and discriminative molecular representation suitable for downstream decision modules.

**3) Decision Tree with Adaptive Feature Weighting.** To improve the accuracy of model decision-making, we incorporate a feature-weighted decision tree at the output stage of the fully connected network. This module dynamically learns feature-specific weights and biases during training, enabling the transformation of high-dimensional molecular representations into probabilistic routing decisions. At each internal node, the tree adaptively adjusts the contribution of individual features, guiding samples along optimized decision paths. The hierarchical weighting mechanism progressively refines feature importance across different levels of the tree, ultimately leading to leaf nodes that produce final predictions through a weighted aggregation of molecular characteristics. This architecture supports both accurate classification and interpretable analysis of feature importance in molecular datasets.

**B. Improvements to HGNN**

As illustrated in Figure S6 (b), we have implemented three key improvements to the conventional HGNN.

**1) Hierarchical Multi-Scale Hypergraph Neural Network**. Traditional Hypergraph Neural Networks typically construct hypergraphs at a single scale (e.g., atomic level), which limits their ability to represent multi-level features such as atoms, residues, and structural domains. Furthermore, the hyperedges in traditional hypergraphs are mainly based on general topological relationships (e.g., spatial distances), lacking biophysical constraints. To address these limitations, we propose an enhanced model—Hierarchical Multi-Scale Hypergraph Neural Network. This model constructs hypergraphs at multiple scales and incorporates biological prior knowledge, enabling a comprehensive modeling of the three-dimensional structure of olfactory receptors. We designed three distinct levels of hypergraph hierarchies. Firstly, Atomic-Level Hypergraph. In this level, atoms are represented as nodes, and hyperedges are constructed based on chemical bonds, hydrogen bonds, hydrophobic interactions, etc. This layer primarily predicts interaction energies at the atomic level. Secondly, Residue-Level Hypergraph. Here, amino acid residues serve as nodes, and hyperedges are constructed based on spatial proximity, capturing the spatial arrangement and functional synergistic relationships between residues. This level is used to identify functional sites and binding residues. Thirdly, Domain-Level Hypergraph. In this layer, functional domains (such as transmembrane helices, loops, etc.) are treated as nodes, and hyperedges are built based on spatial connections and functional coupling between domains. This level reflects the interactions between different functional modules of the receptor and infers the synergistic relationships between receptor structures. This hierarchical design enables the model to learn structural features at different scales, avoiding information loss that may arise from single-scale modeling. Ultimately, the model outputs a receptor-level representation vector by globally integrating features from all levels.

**2) Transformer-Enhanced Receptor Feature Augmentation Strategy.** Building upon the extraction of three-dimensional topological structures of receptors via HGNN, this strategy further incorporates a transformer architecture to fully leverage the biological information embedded within receptor amino acid sequences. This module treats the receptor sequence as a hierarchically semantic biological language, employing multi-head self-attention to model global interactions among all residues and dynamically compute association weights between any pair of amino acid residues. By adaptively allocating attention weights, the module effectively captures long-range dependencies among residues. The feature augmentation strategy enables the model to automatically identify clusters of key residues that, although distributed discretely along the receptor sequence, are spatially proximate in the three-dimensional folded structure and collectively form functional sites, such as ligand-binding pockets. Through the stacking and integration of multiple attention layers, the model progressively focuses on structurally critical regions—active centers and allosteric regulatory sites. Furthermore, transformer-enhanced mechanism simulates the complex residue interaction network within the receptor at the feature-learning level, thereby facilitating a semantically informed reconstruction of the receptor's functional conformation. Finally, the incorporation of the transformer module enhances the model's capacity to represent global sequence features of receptors, and improves its ability to capture cross-sequence synergistic effects and functional–structural correlations. Consequently, it provides more discriminative and biologically interpretable feature representations for downstream tasks.

**3) Adaptive Cross-Modal Attention Fusion Module.** The aforementioned two improvements extract receptor features from three-dimensional geometric topology and one-dimensional sequence semantics. However, simple feature concatenation fails to effectively bridge the semantic gap between one-dimensional sequences and three-dimensional structures. To address this multimodal heterogeneity issue, we proposed an adaptive cross-modal attention fusion mechanism. This module consists of the following two components: Firstly, Cross-Modal Interactive Attention Mechanism. Bidirectional cross-modal attention is introduced to establish dynamic interactions between structural and sequence features. Specifically, the sequence features extracted by the Transformer are used as query vectors to attend to the key-value pairs of structural features extracted by HGNN; conversely, structural features can also serve as queries to interact with sequence features. This mechanism facilitates cross-modal semantic alignment. For instance, when a key residue is identified by the sequence branch, the attention mechanism automatically guides the model to focus on the local chemical environment of that residue in the atomic-level hypergraph (e.g., hydrogen bonding networks, hydrophobic interactions), thereby achieving precise association between sequence features and structural constraints. Secondly, Dynamic Gated Fusion. To address the issue of inconsistent quality in receptor structural data, a dynamic gating network with feature confidence awareness is designed. This network adaptively adjusts the fusion weights of multi-source information based on the inherent uncertainty of each input modality. For example, when structural features are detected to contain noise, the gating unit automatically reduces the contribution weight of the structural branch and enhances the influence of the more reliable sequence features. This design significantly improves the robustness of the model in scenarios with imbalanced data quality. Through the above mechanisms, the model ultimately outputs a high-integrity receptor representation, providing more robust and interpretable multimodal features for downstream tasks.

**C. Enhanced Fully Connected Network Module**

As illustrated in Figure S6 (c), building upon the improved DNDF for extracting molecule structural features and the enhanced HGNN for extracting receptor structural features, we further design the fully connected network module to effectively integrate features from both modalities for predicting molecule-receptor response curves. To further enhance the performance of this module, we incorporate cross-modal attention mechanisms and residual connection-based feature fusion strategies. The specific details are introduced as follows.

**1) Cross-Modal Attention Fusion.** To precisely model the interactions between molecules and receptors, we introduced a feature fusion mechanism incorporating cross-modal attention into the fully connected network. This module first performs bidirectional attention-based modeling between molecular structural features derived from the Deep Neural Decision Forest (DNDF) and the three-dimensional topological features of receptors extracted by the Hypergraph Neural Network (HGNN). Using molecular features as query vectors and receptor features as key-value pairs, the module computes attention weights to enable the model to focus on key binding regions within the receptor structure, such as critical residue clusters within the active site. Furthermore, receptor features can also serve as queries to attend to key functional groups or atomic arrangements in the molecule, establishing bidirectional cross-modal perception and alignment. This mechanism strengthens the semantic association between molecules and receptors in the feature space, and allows the model to adaptively identify structural features closely related to binding affinity, such as hydrogen bonds of specific amino acid residues or aromatic ring characteristics within the molecule. Through attention weight integration, the module reveals critical atomic and residue interaction patterns during molecule-receptor binding, providing a structurally interpretable basis for the model's predictions.

**2) Residual Connections and Gating Feature Fusion.** To further enhance the depth and stability of feature extraction, we incorporate residual connections and a gated multi-scale fusion mechanism into the fully connected network. The module consists of a stack of residual blocks, each containing two fully connected layers with skip connections, effectively mitigating the vanishing gradient problem during deep network training. This enables the model to progressively learn high-order interaction features ranging from atomic-level physicochemical compatibility (e.g., charge complementarity, hydrophobic effects) to residue-level conformational synergy. Concurrently, the network employs a gating mechanism to adaptively fuse feature representations at different levels of abstraction: lower-layer features capture detailed local interactions between atoms, while higher-layer features encode global relationships between the overall receptor conformation and molecular binding patterns, thereby achieving multi-granularity modeling of molecule-receptor interactions. Finally, a constrained output layer maps the fused features to the dynamic parameters of the response curve (amplitude, damping coefficient, frequency, and phase). By incorporating physiological constraints (such as discharge frequency limits and decay time scales), the predicted outcomes are ensured to align with the biological plausibility of neural responses. This design significantly enhances the model's ability to represent complex interaction patterns while improving training efficiency and generalization performance across diverse receptors and molecules.

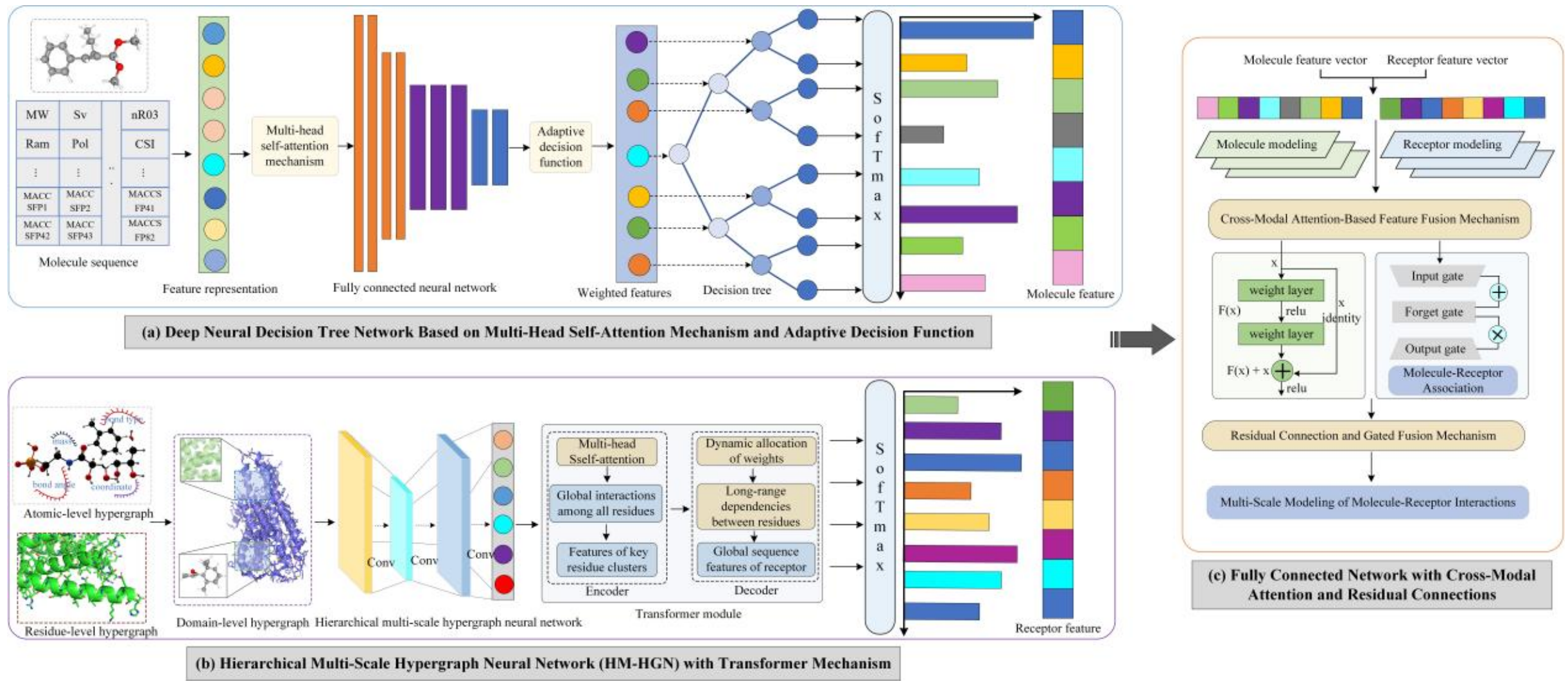

**Fig. S6. Overall Flowchart of the Network Model Improvement**

## V. Model Training and Advanced Models

### A. Model Training

To effectively train the deep neural network model proposed in this study, we adopted a rigorous data partitioning and training protocol. The entire dataset was randomly divided into training, validation, and independent test sets in an 8:1:1 ratio, ensuring that model training, hyperparameter tuning, and final evaluation were all conducted on mutually exclusive data subsets. The validation set was used exclusively for model selection and hyperparameter optimization, whereas the test set was reserved for final performance reporting to prevent information leakage. To further enhance the robustness of model evaluation, we employed 10-fold cross-validation during the training stage. In each fold, the model was trained on 80% of the data and validated on the remaining 20%, ensuring consistent and stable performance across different data partitions. All results reported in this study correspond to the average performance across the 10 folds. Model training was performed using the AdamW optimizer, with an initial learning rate of $1\times10^{-3}$ and a weight decay coefficient of $1\times10^{-4}$. To improve convergence stability and efficiency, we applied a cosine annealing learning rate scheduler, dynamically adjusting the learning rate throughout training. The batch size was set to 128 to balance memory consumption and computational efficiency. The model was trained for 352 epochs, with a total training time of approximately 12 hours. All experiments were implemented using PyTorch 1.11.0 and the Deep Graph Library, and executed on a Linux server equipped with an NVIDIA RTX 2080 GPU (40 GB VRAM).

**B. Advanced Models**

Amit et al. proposed an interpretable metric learning framework based on semantic descriptors. This method builds a structure-to-perception model that predicts intensity, pleasantness, and multiple semantic descriptors directly from molecular structures. The perceptual representation of odor mixtures is then obtained by simply averaging the predicted values of individual components. The method achieved superior root mean square error across multiple datasets. However, this approach relies on a linear averaging strategy, implicitly assuming that the overall perception of a mixture can be represented as a weighted combination of independently predicted single-molecule responses. This assumption neglects the complex nonlinear interactions among molecules, thereby limiting its biological plausibility.

Ravia et al. represented each multi-molecular mixture as a vector composed of physicochemical descriptors and quantified odor similarity by computing angular distances between these vectors. To incorporate concentration effects, the model introduced perceptual intensity as a weighting factor, transforming single-molecule intensity scores into weights via a sigmoid function. These weights were then used during vector aggregation to account for concentration-dependent contributions, enabling a more realistic representation of mixture perception. While this approach demonstrates strong predictive consistency across multiple datasets, it is primarily limited to measuring perceptual similarity between mixtures and does not explicitly predict the specific perceptual attributes of odor mixtures. As a result, it lacks explicit semantic interpretability in terms of detailed odor perception description.

Both POM and MolFormer represent state-of-the-art approaches in the domain of olfactory perception prediction. POM is the current gold standard in olfactory perception prediction by capturing both local and global interactions in molecule structures, which leverages the power of graph neural networks to model complex molecule interactions within a chemical graph. It has proven to be highly accurate for predicting odor perceptions. However, POM is primarily designed for single-molecule odor perception prediction and perceptual space representation. It does not explicitly model the biological transduction process from molecule–receptor interactions to neural encoding, nor does it directly characterize competitive or synergistic interactions among multiple molecules in odor mixtures. In addition, concentration effects are not explicitly incorporated into the POM representation, which limits its ability to model concentration-dependent mixture perception. Therefore, although POM provides a strong structure-to-perception baseline, it remains limited in predicting the perceptual outcomes of multi-molecule mixtures where component composition, concentration ratios, and nonlinear interactions jointly shape the final odor perception.

MolFormer is a large-scale self-supervised molecular language model trained on SMILES sequences using transformer-based architectures with linear attention and rotary positional embeddings. It leverages over one billion unlabeled molecules to learn general-purpose molecular representations that are transferable across diverse downstream property prediction tasks. However, since MolFormer operates on sequence-based SMILES representations rather than explicit molecular graphs or physical interaction structures, it lacks explicit modeling of 3D geometric interactions and interatomic spatial relationships. Moreover, it limits the direct applicability to structured perception modeling tasks.

## VI. Application Areas and Future Work

### A. Application Areas

The proposed method has broad practical applicability across various fields, and can significantly promote the development of relevant industries while reducing trial-and-error costs. The specific applications are elaborated as follows.

#### (1) Embodied Intelligent Robots

The mixture olfactory perception deep learning framework proposed in our study provides embodied intelligent robots with interpretable olfactory perception capabilities, thereby addressing a critical technological gap in the current multimodal perception systems of robotics. Its integration into robotic systems will significantly expand the application depth of robots in complex scenarios such as safety monitoring, healthcare, and environmental interaction, effectively promoting the transformation of machines from passive tools into active collaborative partners [44]. In system integration, this method can be embedded into the robot's perception-decision loop, where it processes signals from gas sensor arrays in real time to output structured and traceable olfactory perception results. These results can be fused with other perceptual modalities—such as vision and hearing—to form a comprehensive environmental cognitive representation. This capability will notably enhance the application potential of robots in high-risk security inspections, non-invasive early disease screening, dynamic environmental pollutant tracking, and personalized human-robot interaction [45]. Rather than being limited to the passive execution of predefined tasks, robots equipped with this framework will be able to actively identify anomalies, predict risks, and adjust behavioral strategies based on real-time olfactory perception. Therefore, our study will advance embodied intelligence toward multimodal active perception and autonomous decision-making.

#### (2) Food and Flavor Industry

Our method has significant application potential in the fields of food science and the flavor and fragrance industry. Traditional flavor formulation development heavily relies on expert experience and time-consuming trial-and-error experiments, which are both costly and inefficient. Given that our model can accurately predict odor perception under different molecule combinations and concentration ratios, it can serve as a powerful computer-aided formulation design tool [46]. For instance, in perfume development, the model can predict the fragrance profiles resulting from the mixing of different ingredients, enabling perfumers to efficiently select candidate formulations and significantly reduce the waste of expensive materials. Furthermore, our method also supports odor reverse engineering. By analyzing the odor perception characteristics of a target product, it can deduce the potential molecule composition and concentration ratios. This capability will accelerate the development of natural flavor substitutes and optimize the flavor of low-sugar foods, offering substantial commercial and research value.

#### (3) Environmental Monitoring

Our method provides a novel solution for environmental monitoring and electronic nose technology. The odors generated in industrial emissions and urban waste treatment are often complex mixtures, and traditional detection methods based on single sensors are typically inadequate for accurately assessing their actual impact on human perception (such as odor intensity or pleasantness). Our model can effectively handle the significant fluctuations in molecule complexity present in real-world environments, enabling real-time gas quality assessment. By integrating this model into intelligent sensing systems, it is possible to establish air quality monitoring systems that better align with human olfactory standards, providing early warnings for chemical plant leaks or environmental odor disturbances [47].

**(4) Digital Olfaction**

The proposed method opens up new avenues for digital olfaction and interactive experiences in virtual reality, offering a solution to address the lack of olfactory perception in human-machine interactions. The core challenge in achieving highly realistic olfactory experiences lies in synthesizing a virtually infinite variety of target odors from a limited set of scent primitives [48]. Given that our model successfully constructs an accurate end-to-end mapping from molecule structure to neural response to odor perception, it can serve as the core computational module for virtual olfactory generators. Specifically, by defining the perceptual descriptors of the target odor, the model can calculate and optimize the mixing ratios of the odor primitives in the scent generator to minimize the perceptual difference between the synthesized and target odors in the perceptual space. This capability enables scent transmission, where the sender transmits the digital olfactory perception code, and the receiver uses our method to dynamically generate the scent in real-time [49]. This not only facilitates high-fidelity odor reproduction in remote presence applications but also holds significant potential for immersive gaming, remote medical diagnosis.

**B. Future Work**

Although our method has demonstrated superior performance in accurately identifying the odor perception of multi-molecule mixtures, further optimization is still required. Future work will focus on the following areas:

**(1) Enhancement of Model Explainability**

Although the deep learning-based method proposed in this study achieves high-precision perception prediction, the internal decision-making logic of deep neural networks remains opaque. Future work will focus on enhancing the explainability of the model. Specifically, we plan to analyze the distribution of attention weights in the model, with an emphasis on identifying the molecule regions the model focuses on when processing complex mixtures. We aim to establish a mapping mechanism that links specific molecule substructures or pharmacophores to distinct neural response patterns. More importantly, by comparing the attention score distributions between single molecules and mixtures, we hope to capture the nonlinear interaction rules that lead to odor masking or synergy effects. This will help reveal the combinatorial coding logic of the olfactory system from a data-driven perspective and discover novel biochemical mechanisms of olfaction.

**(2) Expansion of Personalized Perception**

Our method demonstrated excellent performance in predicting the odor perception of the general population, but it has yet to account for individual differences in perception. The polymorphisms in human olfactory receptor genes lead to significant variations in the way different individuals perceive the same odorant molecules or mixtures. These differences are not only related to an individual's genetic background but also to demographic characteristics such as age and gender.

Future work will focus on integrating individual genomic data with demographic features to advance the development of a personalized olfactory model. We aim to more accurately predict the olfactory preferences of specific populations or individuals. This effort will provide a scientific foundation for personalized health and perfume customization, driving the application potential of olfactory perception.

**(3) Multimodal Feature Fusion**

To build a more advanced machine olfaction system, we will explore the potential of multimodal learning, particularly in integrating neurobiological signals. Specifically, we plan to combine molecule structures with electroencephalography and functional magnetic resonance imaging signals, using multimodal learning approaches to establish an end-to-end mapping from molecule stimuli to cortical responses. This will enable a deeper understanding of how the brain perceives and processes mixed odor information, particularly the correlation between neural activity and odor perception. By integrating multimodal signals, we can develop more accurate models to simulate the brain's processing of odors, further enhancing the performance of machine olfaction systems. Additionally, this research offers a new direction for the application of brain-computer interfaces in the olfactory domain. By integrating odor perception models, it will be possible to achieve more precise perceptual interactions, advancing fields such as intelligent health monitoring and virtual olfaction.

## References

[1] ODORactor: a web server for deciphering olfactory coding. Bioinformatics (Oxford, England) (2011).

[2] Modena, D. , et al. OlfactionDB: A Database of Olfactory Receptors and Their Ligands. Advances in Life Sciences 1.1(2011):1-5. DOI:

[3] Achebouche R, Tromelin A, Audouze K, et al. Application of artificial intelligence to decode the relationships between smell, olfactory receptors and small molecules[J]. Scientific reports, 2022, 12(1): 18817. DOI: https://doi.org/10.1038/s41598-022-23176-y

[4] The Good Scents Company. (2024). *The Good Scents Company Information System*. Retrieved from

[5] Lefngwell & Associates. Flavor-Base. 9th Edition. Available online: http://www.lefngwell.com/ favbase.htm

[6] Snitz K, Yablonka A, Weiss T, et al. Predicting odor perceptual similarity from odor structure[J]. PLoS computational biology, 2013, 9(9): e1003184.

[7] Amit Dhurandhar, Hongyang Li, Guillermo A Cecchi, Pablo Meyer, Expansive linguistic representations to predict interpretable odor mixture discriminability, *Chemical Senses*, Volume 48, 2023, bjad018.

[8] Kowalewski J, Ray A. Predicting human olfactory perception from activities of odorant receptors[J]. IScience, 2020, 23(8).

[9] Tisserand R, Young R. Essential oil safety: a guide for health care professionals[M]. Elsevier Health Sciences, 2013.

[10] Reisert J, Matthews H R. Adaptation of the odour-induced response in frog olfactory receptor cells[J]. The Journal of physiology, 1999, 519(3): 801-813.

[11] Rokni D, Hemmelder V, Kapoor V, et al. An olfactory cocktail party: figure-ground segregation of odorants in rodents[J]. Nature neuroscience, 2014, 17(9): 1225-1232.

[12] Nagel K I, Wilson R I. Biophysical mechanisms underlying olfactory receptor neuron dynamics[J]. Nature neuroscience, 2011, 14(2): 208-216.

[13] Rospars J P, Lánský P, Duchamp-Viret P, et al. Characterizing and modeling concentration-response curves of olfactory receptor cells[J]. Neurocomputing, 2001, 38: 319-325.

[14] Bathellier B, Buhl D L, Accolla R, et al. Dynamic ensemble odor coding in the mammalian olfactory bulb: sensory information at different timescales[J]. Neuron, 2008, 57(4): 586-598.

[15] Friedrich R W, Laurent G. Dynamic optimization of odor representations by slow temporal patterning of mitral cell activity[J]. Science, 2001, 291(5505): 889-894.

[16] Stopfer M, Bhagavan S, Smith B H, et al. Impaired odour discrimination on desynchronization of odour-encoding neural assemblies[J]. Nature, 1997, 390(6655): 70-74.

[17] Kurahashi T, Menini A. Mechanism of odorant adaptation in the olfactory receptor cell[J]. Nature, 1997, 385(6618): 725-729.

[18] Lapid H, Shushan S, Plotkin A, et al. Neural activity at the human olfactory epithelium reflects olfactory perception[J]. Nature neuroscience, 2011, 14(11): 1455-1461.

[19] Buzsaki G, Draguhn A. Neuronal oscillations in cortical networks[J]. science, 2004, 304(5679): 1926-1929.

[20] Mori K, Nagao H, Yoshihara Y. The olfactory bulb: coding and processing of odor molecule information[J]. Science, 1999, 286(5440): 711-715.

[21] Laurent G. Olfactory network dynamics and the coding of multidimensional signals[J]. Nature reviews neuroscience, 2002, 3(11): 884-895.

[22] Kay L M, Beshel J, Brea J, et al. Olfactory oscillations: the what, how and what for[J]. Trends in neurosciences, 2009, 32(4): 207-214.

[23] Soucy E R, Albeanu D F, Fantana A L, et al. Precision and diversity in an odor map on the olfactory bulb[J]. Nature neuroscience, 2009, 12(2): 210-220.

[24] Ni Y, Feng S, Hong X, et al. Pre-training with fractional denoising to enhance molecular property prediction[J]. Nature Machine Intelligence, 2024, 6(10): 1169-1178.

[25] Wachowiak M, Cohen L B. Representation of odorants by receptor neuron input to the mouse olfactory bulb[J]. Neuron, 2001, 32(4): 723-735.

[26] Reisert J, Zhao H. Response kinetics of olfactory receptor neurons and the implications in olfactory coding[J]. Journal of General Physiology, 2011, 138(3): 303-310.

[27] Singh V, Murphy N R, Balasubramanian V, et al. Competitive binding predicts nonlinear responses of olfactory receptors to complex mixtures[J]. Proceedings of the National Academy of Sciences, 2019, 116(19): 9598-9603.

[28] Carey R M, Verhagen J V, Wesson D W, et al. Temporal structure of receptor neuron input to the olfactory bulb imaged in behaving rats[J]. Journal of neurophysiology, 2009, 101(2): 1073-1088.

[29] Mainland J D, Keller A, Li Y R, et al. The missense of smell: functional variability in the human odorant receptor repertoire[J]. Nature neuroscience, 2014, 17(1): 114-120.

[30] Araneda R C, Kini A D, Firestein S. The molecular receptive range of an odorant receptor[J]. Nature neuroscience, 2000, 3(12): 1248-1255.

[31] Chmiela S, Sauceda H E, Müller K R, et al. Towards exact molecular dynamics simulations with machine-learned force fields[J]. Nature communications, 2018, 9(1): 3887.

[32] Moriwaki H, Tian Y S, Kawashita N, et al. Mordred: a molecular descriptor calculator[J]. Journal of cheminformatics, 2018, 10(1): 4.

[33] Konda R , Reddy S T , Moon S A ,et al. AI-Driven Drug Discovery: Leveraging Machine Learning for Predictive Molecular Design and Accelerated Pharmaceutical Innovation[C]//2025 IEEE 4th World Conference on Applied Intelligence and Computing (AIC).0[2026-01-22].

[34] Shaw D E, Maragakis P, Lindorff-Larsen K, et al. Atomic-level characterization of the structural dynamics of proteins[J]. Science, 2010, 330(6002): 341-346.

[35] Billesbølle C B, de March C A, van der Velden W J C, et al. Structural basis of odorant recognition by a human odorant receptor[J]. Nature, 2023, 615(7953): 742-749.

[36] Kim W K, Choi K, Hyeon C, et al. General chemical reaction network theory for olfactory sensing based on g-protein-coupled receptors: Elucidation of odorant mixture effects and agonist–synergist threshold[J]. The Journal of Physical Chemistry Letters, 2023, 14(38): 8412-8420.

[37] Nagel K I, Wilson R I. Biophysical mechanisms underlying olfactory receptor neuron dynamics[J]. Nature neuroscience, 2011, 14(2): 208-216.

[38] Su C Y, Menuz K, Reisert J, et al. Non-synaptic inhibition between grouped neurons in an olfactory circuit[J]. Nature, 2012, 492(7427): 66-71.

[39] Stopfer M, Jayaraman V, Laurent G. Intensity versus identity coding in an olfactory system[J]. Neuron, 2003, 39(6): 991-1004.

[40] Si G, Kanwal J K, Hu Y, et al. Structured odorant response patterns across a complete olfactory receptor neuron population[J]. Neuron, 2019, 101(5): 950-962. e7.

[41] Chong E, Moroni M, Wilson C, et al. Manipulating synthetic optogenetic odors reveals the coding logic of olfactory perception[J]. Science, 2020, 368(6497): eaba2357.

[42] Singh V, Murphy N R, Balasubramanian V, et al. Competitive binding predicts nonlinear responses of olfactory receptors to complex mixtures[J]. Proceedings of the National Academy of Sciences, 2019, 116(19): 9598-9603.

[43] Pashkovski S L, Iurilli G, Brann D, et al. Structure and flexibility in cortical representations of odour space[J]. Nature, 2020, 583(7815): 253-258.

[44] Chuntae Kim, Kyung Kwan Lee, Moon Sung Kang, Dong-Myeong Shin, Jin-Woo Oh, Chang-Soo Lee, Dong-Wook Han. Artificial olfactory sensor technology that mimics the olfactory mechanism: a comprehensive review. *Biomater Res.* 2022;26:40.

[45] Sun F, Chen R, Ji T, et al. A Comprehensive Survey on Embodied Intelligence: Advancements, Challenges, and Future Perspectives. *CAAI Artificial Intelligence Research*, 2024, 3: 9150042.

[46] Sanchez-Lengeling, B. et al. Machine learning for scent: Learning generalizable perceptual representations of small molecules. *ACS Central Science* 5, 1524–1534 (2019).

[47] Hongyang L , Bharat P , Omenn G S ,et al. Accurate prediction of personalized olfactory perception from large-scale chemoinformatic features[J].GigaScience, 2018(2):1-11.DOI:10.1093/gigascience/gix127.

[48] Yuansheng Zhou *et al.*, Hyperbolic geometry of the olfactory space.*Sci. Adv.***4**,eaaq1458(2018).

[49] Oh J H .A Pattern Recognition Artificial Olfactory System Based on Human Olfactory Receptors and Organic Synaptic Devices[J].Proceedings of the MATSUS Spring 2024 Conference, 2023.DOI:10.29363/nanoge.matsus.2024.013.